\documentclass[]{alaya}
\usepackage{makecell}
\usepackage{wrapfig}
\usepackage{tabularx}
\usepackage{textcomp}
\usepackage{stfloats}
\usepackage{url}
\usepackage{verbatim}
\usepackage{titlesec}
\usepackage{tocloft}
\usepackage{adjustbox}
\usepackage{multirow}
\usepackage{pifont}
\usepackage[sc]{mathpazo}
\usepackage{tikz}
\usepackage{comment}
\usepackage{amsmath,amssymb}
\usepackage{colortbl}
\usepackage{natbib}
\usepackage{color}
\usepackage{booktabs} 
\usepackage{hyperref}
\usepackage{graphicx}
\usepackage{subcaption}
\RequirePackage{xspace}
\makeatletter
\DeclareRobustCommand\onedot{\futurelet\@let@token\@onedot}
\def\@onedot{\ifx\@let@token.\else.\null\fi\xspace}
\usepackage[most]{tcolorbox}
\usepackage{array}
\usepackage{siunitx}
\usepackage[table]{xcolor}
\usepackage{caption}
\definecolor{headerpurple}{HTML}{d8d2fc}
\definecolor{rowgray}{gray}{0.95}
\usepackage{CJKutf8}
\def\eg{\emph{e.g}\onedot}

\makeatother

\definecolor{adptorange}{RGB}{248, 205, 172}
\definecolor{cmpblue}{RGB}{189, 215, 238}

\definecolor{our_red}{RGB}{232,157,160}
\definecolor{our_blue}{RGB}{136,206,230}
\definecolor{our_orange}{RGB}{246,200,168}
\definecolor{our_green}{RGB}{178,211,164}

\definecolor{attn_code0}{RGB}{247,215,200}
\definecolor{attn_code1}{RGB}{238,169,139}
\definecolor{mlp_code0}{RGB}{204,201,221}
\definecolor{mlp_code1}{RGB}{102,95,153}
\definecolor{mygray}{HTML}{f0f0f0}

\definecolor{token_blue}{RGB}{84, 120, 140}

\usepackage{bbding}
\usepackage{fontawesome}
\usepackage{float}

\newcommand{\cmark}{\textcolor{green!70!black}{\ding{51}}}
\newcommand{\xmark}{\textcolor{red!70!black}{\ding{55}}}
\newcommand{\pmark}{\textcolor{orange!80!black}{$\boldsymbol{\sim}$}}

\newlength\savewidth

\newcolumntype{x}[1]{>{\centering\arraybackslash}p{#1pt}}
\newcolumntype{y}[1]{>{\raggedright\arraybackslash}p{#1pt}}
\newcolumntype{z}[1]{>{\raggedleft\arraybackslash}p{#1pt}}

\renewcommand{\paragraph}[1]{\vspace{1.25mm}\noindent\textbf{#1}}

\usepackage{algorithm}
\usepackage{listings}

\definecolor{codeblue}{rgb}{0.25, 0.5, 0.5}
\definecolor{codekw}{rgb}{0.35, 0.35, 0.75}
\lstdefinestyle{Pytorch}{
    language = Python,
    backgroundcolor = \color{white},
    basicstyle = \fontsize{9pt}{8pt}\selectfont\ttfamily\bfseries,
    columns = fullflexible,
    aboveskip=1pt,
    belowskip=1pt,
    breaklines = true,
    captionpos = b,
    commentstyle = \color{codeblue},
    keywordstyle = \color{codekw},
}

\definecolor{green}{HTML}{009000}
\definecolor{red}{HTML}{ea4335}

\newcommand{\method}{\texttt{WorldMark}\xspace}
\newcommand{\figref}[1]{\textcolor{refblue}{Figure~\ref{#1}}}
\newcommand{\tabref}[1]{\textcolor{refblue}{Table~\ref{#1}}}
\newcommand{\secref}[1]{Section~\ref{#1}}

\definecolor{refblue}{HTML}{2A039B}

\hypersetup{colorlinks=true, linkcolor=refblue, citecolor=refblue, urlcolor=refblue}

\title{WorldMark: A Unified Benchmark Suite for Interactive Video World Models}
\author[1,2]{Xiaojie Xu}
\author[1,2]{Zhengyuan Lin}
\author[1,3]{Kang He}
\author[1,3]{Yukang Feng}
\author[1]{Xiaofeng Mao}
\author[1,3]{Yuanyang Yin}
\author[\dagger 1]{Yongtao Ge}
\author[\dagger 1,3]{Kaipeng Zhang}

\affiliation[1]{Alaya Lab}
\affiliation[2]{The University of Tokyo}
\affiliation[3]{Shanghai Innovation Institute}
\abstract{
Unlike text- or image-driven video generation, an interactive world model is driven by actions: the user acts, and the world responds. Two obstacles stand in the way of fair and comprehensive evaluation. First, models take actions in incompatible formats---captions, camera trajectories, action functions---so no shared protocol has been established. Second, while existing benchmarks have advanced world memory and visual quality, action following is reduced to trajectory or direction error, which collapses a whole path into one number: not how quickly the world reacts to a command switch, nor how cleanly it moves along the commanded axis. \method removes both obstacles. Per-model adapters translate a shared WASD-style vocabulary into each model's native control format, so ten heterogeneous models receive semantically identical instructions across 500 standardized cases spanning styles, viewpoints, and difficulty tiers; a new model costs one adapter. On this common ground we characterize action dynamics through a control-systems lens---direction accuracy, direction purity, response latency, and motion stability, each resolved per axis---alongside suites for world memory and visual quality. Together they expose differences existing protocols cannot see: the fastest responders are often the least stable, a trade-off no single action metric captures; per-axis resolution reveals models that follow translation almost perfectly while barely responding to rotation; the model with the best perceptual and aesthetic quality ranks last in translational direction accuracy and latency; and stylized scenes cost every model global consistency while leaving action dynamics largely intact. We will release all data, evaluation code, and model outputs. \textbf{Beyond offline metrics, we launch World Model Arena (\url{warena.ai}), an online platform where anyone can pit leading world models against each other in side-by-side battles and watch the live leaderboard.}
}

\github{\url{https://alayalab.github.io/WorldMark/}}
\metadata[Github Repo]{\url{https://github.com/AlayaLab/WorldMark}}
\metadata[World Model Arena]{\url{https://warena.ai/}}
\date{\today}

\begin{document}
\maketitle
\setcounter{footnote}{1}
\footnotetext{$^\dagger$Corresponding authors: \texttt{yongtao.ge@shanda.com}, \texttt{kaipeng.zhang@shanda.com}}

\section{Introduction}
\begin{figure}[t]
    \centering
    \includegraphics[width=0.55\linewidth]{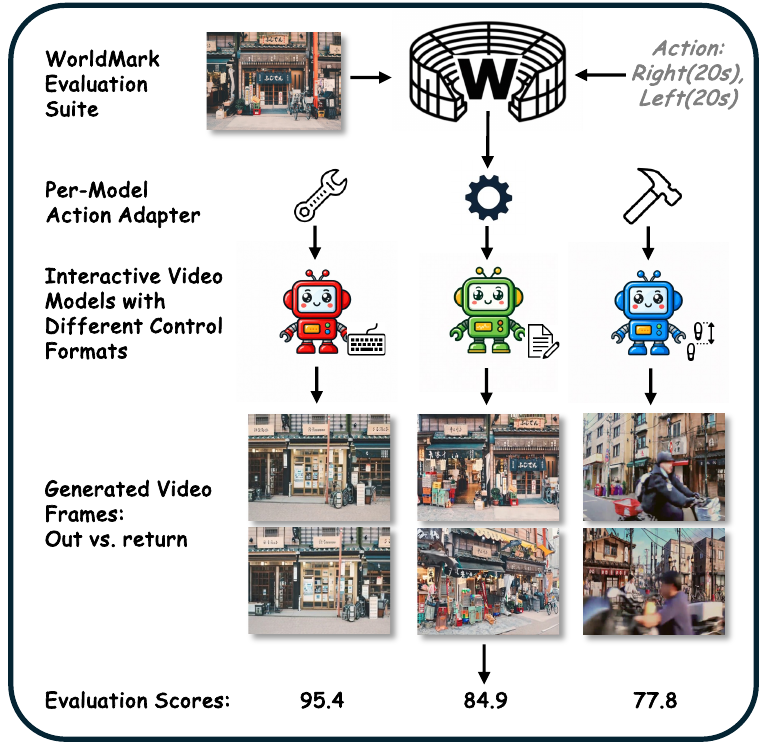}
    \caption{Overview of \method. Models take actions in incompatible formats---keyboard actions, text prompts, camera trajectories---so per-model adapters translate one shared vocabulary into each native format. For a frame on the outbound leg and its return-leg counterpart at equal accumulated motion, the two views should match but diverge to different degrees. We show revisit memory scores as one illustrative axis; the same probe additionally yields action dynamics metrics, and visual quality is scored on the entire video.}
    \label{fig:overview}
    
\end{figure}

Press \texttt{W}, and the world should move forward. Turn away and turn back, and the street you left should still be there. This is what separates an interactive world model~\cite{ha2018world} from text- or image-driven video generation~\cite{zhao2025drivedreamer,cen2025worldvla}: the user is inside the loop. The past few years have produced a wave of interactive video world models~\cite{mao2025yume,he2025matrix,shen2026lyra,team2026alayaworld,team2026dreamx,zhu2026sana} that generate at steadily improving quality, maintain persistent contextual memory, and accept user actions as input.

\begin{table}[t]
  \centering
  \caption{Comparison of evaluation benchmarks along data coverage and metric capability.
           The \emph{Action Dynamics} column requires \emph{response} dynamics
           (latency, persistence, stability)---beyond direction or trajectory accuracy,
           which single-action, no-switch protocols cannot measure in principle.
           The \emph{Full Reproducibility} criterion is bit-identical outputs across repeated runs.
           \cmark~= supported, \pmark~= partial, \xmark~= not supported.
           $^{a}$\,Duration-based (equal-time) frame matching, which misaligns under response latency and gain asymmetry.
           $^{b}$\,No explicit revisit protocol; round-trip cases identified post hoc via SLAM.
           $^{c}$\,Partially relies on SLAM-based estimators with bundle adjustment variance.
           $^{d}$\,Partially relies on generative VLM scoring with sampling variance.}
  \label{tab:benchmark-comparison}
  \setlength{\tabcolsep}{6pt}
  \renewcommand{\arraystretch}{1.15}
  \resizebox{0.92\textwidth}{!}{%
  \begin{tabular}{l l ccc ccc}
    \toprule
    \multirow{2}{*}{\textbf{Benchmark}}
      & \multirow{2}{*}{\textbf{Evaluated Tasks}}
      & \multicolumn{3}{c}{\textbf{Data}}
      & \multicolumn{3}{c}{\textbf{Metrics}} \\
    \cmidrule(lr){3-5} \cmidrule(lr){6-8}
      &
      & \makecell{Diverse\\Styles}
      & \makecell{Diverse\\Viewpoints}
      & \makecell{Difficulty\\Hierarchy}
      & \makecell{Action\\Dynamics}
      & \makecell{Revisit\\Memory}
      & \makecell{Full\\Reproducibility} \\
    \midrule
    VBench~\cite{huang2024vbench, huang2024vbench++}          & T2V, I2V                    & \xmark & \xmark & \xmark & \xmark & \xmark & \cmark \\
    WorldScore~\cite{duan2025worldscore}   & I2V, I2-3D  & \cmark & \cmark & \xmark & \xmark & \xmark & \pmark$^{c}$ \\
    MIND~\cite{mind}                       & I2V, V2V    & \cmark & \cmark & \xmark & \xmark & \cmark & \pmark$^{c}$ \\
    iWorld-Bench~\cite{fang2026iworld}        & Interactive I2V        & \xmark & \cmark & \cmark & \xmark & \pmark$^{a}$ & \pmark$^{c}$ \\
    MBench~\cite{zhang2026mbench}                   & T2V, Interactive I2V   & \xmark & \xmark & \xmark & \xmark & \cmark & \pmark$^{d}$ \\
    WBench~\cite{ying2026wbench}                   & T2V, Interactive I2V   & \cmark & \cmark & \xmark & \xmark & \pmark$^{b}$ & \pmark$^{d}$ \\
    \midrule
    \textbf{\method (Ours)}                & \textbf{Interactive I2V} & \cmark & \cmark & \cmark & \cmark & \cmark & \cmark \\
    \bottomrule
  \end{tabular}%
  }
\end{table}

However, benchmarks for these interactive systems still concentrate on visual quality and world memory, and largely overlook how the world responds to what the user does. Quality-oriented suites~\cite{huang2024vbench,huang2024vbench++} treat video as something to be watched rather than driven. Trajectory-conditioned benchmarks~\cite{duan2025worldscore,lu20254dworldbench} brought 3D-aware consistency, but a prescribed camera path substitutes for interaction rather than testing it. And benchmarks designed for world models specifically~\cite{mind,zhang2026mbench,ying2026wbench,fang2026iworld} have promoted memory to a first-class concern, asking whether a scene stays consistent when revisited. \textbf{Visual quality and memory are therefore well served; action following is not.} Two obstacles have kept it there. First, models receive instructions in different languages---keyboard tokens embedded in captions~\cite{mao2025yume}, 6-DoF pose parameters~\cite{team2025hunyuanworld}, camera trajectories~\cite{shen2026lyra}, custom action functions~\cite{he2025matrix}---so it is difficult to drive different models with identical instructions on the same scene, and some benchmarks sidestep this by conditioning on ground-truth camera trajectories, which most keyboard-controlled models cannot accept. Second, current evaluation only focuses on trajectory error or direction accuracy, collapsing an entire path into a single number that quantifies how far a model deviates without revealing what it did wrong.

Watching a large number of generated videos, we repeatedly encountered situations a simple error metric cannot distinguish. Told to move forward, some models drift steadily off-axis, curving away while their net displacement stays forward. Told to stop turning and walk straight, some keep turning for seconds after the command changed. Told to keep going, some models stutter, stall, or slow to a standstill before the command ends. To anyone at the controls, these differences separate a world that obeys from one that disobeys.

We therefore ask: \textbf{can we build a unified benchmark that accommodates every interactive control format, spans visual styles, viewpoints, and difficulty levels, and whose results expose the problems above?} We argue that this requires two things. First, every model must execute the same actions on the same scenes through its own native interface---otherwise responses cannot be compared. Second, response needs a vocabulary, and control theory~\cite{doyle2013feedback} already supplies one: a system under a step input is characterized not by its final position but by how it gets there---settling time, overshoot, steady-state error. \textbf{An action command is a ``step input'', and a world model's response to one can be described the same way.}

\method builds both. A shared vocabulary of six primitives (WASD for translation, L/R for yaw) is translated into each model's native format by a per-model adapter, so that ten heterogeneous models receive semantically identical instructions at their own native motion scale; adding a model costs one adapter. These instructions are issued over a standardized suite of 500 cases---50 scenes in real and stylized domains, each in paired first-/third-person views, the resulting 100 images each issued five of 15 trajectories at three difficulty tiers. On this common ground, four Action Dynamics metrics---direction accuracy, direction purity, response latency, and motion stability, each resolved separately for translation and rotation---target exactly the failures above, while three World Memory metrics separate the local, global, and revisit timescales and two Visual Quality metrics complete the picture. All metrics use feed-forward estimators only, producing bit-for-bit deterministic results across runs.

In summary, our contributions are as follows:
\begin{itemize}
    \item \textbf{We introduce a unified action interface} that drives ten models with incompatible control formats from a single action vocabulary, together with a standardized suite of 500 cases spanning visual styles, viewpoints, and difficulty tiers; supporting a new model requires one adapter.
    \item \textbf{We propose an action-centric metric suite} that treats each command as a step input, after control-systems practice, and characterizes \emph{how} a world responds---when motion begins, whether it stays on the commanded axis, and whether it holds---rather than only how far it strays, resolved per motion axis and complemented by memory metrics at three timescales and two visual quality scores.
    \item \textbf{We conduct a systematic evaluation} that turns these measurements into actionable findings: a response--stability trade-off no single action metric can capture, command axes that fail independently, and a negative correlation between perceptual quality and responsiveness that contradicts appearance-based evaluation.
\end{itemize}

\section{Related Works}
\subsection{Interactive Video World Models}
\label{sec:related_work_world_simulation}

Video generation has advanced from controllable diffusion architectures~\cite{ho2022video,he2024cameractrl} to large-scale transformers~\cite{liu2024sora,yangcogvideox,kong2024hunyuanvideo,wan2025wan}, prompting researchers to treat video generators as interactive world simulators~\cite{yang2023unisim}. This was first realized within the Minecraft domain~\cite{guo2025mineworld,oasis2024}, establishing real-time action-conditioned generation but confined to a single environment.

The current generation reaches open photorealistic and stylized worlds through markedly different control interfaces: Yume~1.5~\cite{mao2025yume} embeds directional keywords in captions; HY-World~1.5~\cite{team2025hunyuanworld} and HY-GameCraft~\cite{li2025hunyuan,tang2025hunyuan} take 6-DoF pose parameters; Lyra~2.0~\cite{shen2026lyra} takes camera trajectories; Matrix-Game~2.0~\cite{he2025matrix} and 3.0~\cite{wang2026matrix} expose discrete action functions with streaming inference, as does WorldPlay~\cite{sun2025worldplay}; AlayaWorld~\cite{team2026alayaworld}, DreamX-World~\cite{team2026dreamx}, LingBot-World~\cite{team2026advancing}, and SANA-WM~\cite{zhu2026sana} each define their own action space; and Genie~3~\cite{deepmind2025genie3} maps gamepad controls behind a closed interface. This heterogeneity fragments evaluation: each system reports on a private benchmark with its own scenes and metrics---yume-Bench~\cite{mao2025yume}, GameWorld Score~\cite{zhang2025matrix}, InterBench~\cite{tang2025hunyuan}---so published numbers cannot be compared, and no protocol reaches all models without a translation layer.

\subsection{Benchmarks for Video World Models}
\label{sec:related_work_benchmark}

General-purpose benchmarks established the measurement vocabulary the field still uses: VBench and VBench++ ~\cite{huang2024vbench, huang2024vbench++} decompose generation quality into multiple dimensions, while PhyGenBench~\cite{meng2025towards} and WorldModelBench~\cite{li2025worldmodelbench} probe physical plausibility---none with a user in the loop. Another family conditions on camera motion: WorldScore~\cite{duan2025worldscore} and 4DWorldBench~\cite{lu20254dworldbench} prescribe trajectories and score the resulting geometry through 3D/4D reconstruction. Such measurements are 3D-aware, but the camera path is supplied rather than requested, an input interface most keyboard-controlled models do not expose.

Concurrent benchmarks target interactive models directly, each handling heterogeneous action interfaces differently. MIND~\cite{mind} requires ground-truth camera poses as input, restricting it to models with continuous pose control. MBench~\cite{zhang2026mbench} matches revisits by estimated pose, but splits text- and action-conditioned models into tracks sharing no common cases, leaving same-named scores incomparable. WBench~\cite{ying2026wbench} and iWorld-Bench~\cite{fang2026iworld} come closest, both re-expressing one action vocabulary as text, pose, or keys; yet unification covers only part of their suites, so shared comparison collapses to a navigation subset. Three limitations recur on the action side. First, action following is reported as direction accuracy or trajectory error, single scalars that leave drift, delay, and stalling unresolved and never separate motion axes. Second, revisit probes sit outside the action protocol---anchored to ground-truth poses, or recovered afterwards by elapsed-time or pose matching: the former misaligns whenever a model responds late or returns at a different speed than it departed, the latter inherits its estimator's uncertainty. Third, geometric scores come from SLAM back-ends~\cite{Droid-slam,huang2025vipe}, whose bundle adjustment is not deterministic, while other dimensions go to generative VLM judges, whose sampled outputs vary---neither reproduces across runs.

\method addresses all three (\tabref{tab:benchmark-comparison}): it issues identical action sequences to ten models through per-model adapters rather than requiring a shared input modality, and every metric operates on generated video alone, so no dimension is confined to a subset of models; it decomposes action following into when a response begins, whether it stays on the commanded axis, and whether it persists, each reported separately for translation and rotation; and its revisit probe is built into the action protocol, pairing frames by equal accumulated motion without ground-truth frames or reconstruction. All metrics use feed-forward estimators alone and reproduce bit-for-bit across runs.
\section{WorldMark Suite}

\method comprises five components: (i)~an \textbf{Evaluation Dimension Suite} of nine metrics spanning Action Dynamics, World Memory, and Visual Quality; (ii)~an \textbf{Image Suite} covering diverse scenes, styles, and viewpoints; (iii)~an \textbf{Action Suite} of 15 trajectories organized into difficulty tiers; (iv)~a \textbf{Unified Action Interface} that translates a shared WASD+L/R vocabulary into each model's native control format; and (v)~a modular \textbf{Evaluation Workflow} tying these into a reproducible run. Users may also apply custom metrics to our standardized image--action pairs for fair evaluation.

\begin{table}[htbp]
\centering
\caption{The nine evaluation dimensions of \method.}
\label{tab:eval_dimensions}
\setlength{\tabcolsep}{5pt}
\renewcommand{\arraystretch}{1.15}
\resizebox{0.5\linewidth}{!}{%
\begin{tabular}{l l}
\toprule
\textbf{Category} & \textbf{Metrics} \\
\midrule
Action Dynamics & \makecell[l]{Direction Accuracy; Direction Purity;\\Response Latency; Motion Stability} \\
\addlinespace[2pt]
World Memory    & \makecell[l]{Local; Global; Revisit Memory} \\
\addlinespace[2pt]
Visual Quality  & Perceptual; Aesthetic Quality \\
\bottomrule
\end{tabular}%
}
\end{table}

\subsection{Evaluation Dimension Suite}
\label{sec:sec_3.1}

\tabref{tab:eval_dimensions} lists our metrics in three categories. \textbf{Action Dynamics} characterizes how motion reacts to each command; \textbf{World Memory} asks whether the generated world stays the same, at three separate scales; \textbf{Visual Quality} measures per-frame fidelity, testing whether the other two reduce to appearance. All thresholds are fixed across models and splits; the supplement gives their detailed values and a sensitivity analysis.

\subsubsection{Action Dynamics}

An action command is a step input, yet a trajectory error integrates the whole response into one number: a model that drifts off-axis, reacts late, or stalls midway can score like a model that does none of these. We instead ask four questions: \textbf{does motion go the right way, only that way, soon enough, and does it hold steadily?}

Answering them from camera poses fails: both SLAM~\cite{Droid-slam,huang2025vipe} and feed-forward pose estimators~\cite{lin2025depth} assume a rigid scene, exactly what a failing generative model cannot produce. Displacement is also not comparable across models moving at different speeds. Instead, optical flow~\cite{wang2024sea} needs no reconstruction, and reading it through ratios keeps the metrics scale-free. We therefore take the measured flow as our proxy for camera motion.

Each video frame reduces to a horizontal component $S_x$ and a radial component $S_{\text{rad}}$ about the focus of expansion, matched against the template expected for the commanded key---radial for W/S, horizontal otherwise. Strafing and yawing share that template, so depth~\cite{lin2025depth} separates them by parallax: translation moves near content more than far, rotation moves everything alike. Under small motion, the horizontal flow $u$ at a pixel of depth $Z$ therefore splits into
\begin{equation}
u = \underbrace{f t_x}_{a}(1/Z) \underbrace{- f \omega_y}_{b}, \qquad
p_{\text{tr}} = \frac{|a|\,\Delta(1/Z)}{|a|\,\Delta(1/Z) + |b|},
\label{eq:depth_split}
\end{equation}
for focal length $f$, lateral velocity $t_x$, yaw rate $\omega_y$, and $\Delta(1/Z)$ the inverse-depth span. We fit $a,b$ by least squares over sampled $(1/Z,u)$ pairs; $f$ cancels in $p_{\text{tr}}$, leaving the split free of intrinsics that differ across models.

\noindent\textbf{Direction Accuracy and Purity.} Accuracy asks whether the world moves the commanded way, purity whether it moves \emph{only} that way: strafing left while creeping forward scores well on the first, poorly on the second. With $v$ the commanded channel signed by its expected direction and $E_{\text{lat}}=p_{\text{tr}}|S_x|$, $E_{\text{yaw}}=(1-p_{\text{tr}})|S_x|$, $E_{\text{fwd}}=|S_{\text{rad}}|$ the three channels,
\begin{equation}
\mathrm{DA} = \frac{\overline{v}}{\overline{|v|}}\cdot w, \qquad
\mathrm{DP} = \frac{E_{\text{cmd}}}{E_{\text{lat}}+E_{\text{yaw}}+E_{\text{fwd}}},
\label{eq:da_dp}
\end{equation}
where $w\!\in\!\{p_{\text{tr}}, 1\!-\!p_{\text{tr}}, 1\}$ gates by motion type, so turning when asked to strafe scores low despite correct flow.

\noindent\textbf{Response Latency.} A model still executing the previous command after the input changes is not responsive. We time how long the new command takes to reach half the speed the model normally sustains, normalized by segment length $L$:
\begin{equation}
\tau = \min\{t : \tilde v(t) \ge \tfrac{1}{2} v_{\text{ref}}\} - t_0, \qquad
\mathrm{RL} = 1 - \frac{\min(\tau, L)}{L},
\label{eq:latency}
\end{equation}
with $\tilde v$ a $0.5$\,s moving average and $v_{\text{ref}}$ the $75$th percentile of channel magnitude. A global rather than per-segment reference stops a weak responder lowering its own bar; segments never reaching it, or below $0.1\,v_{\text{ref}}$, score 0.

\noindent\textbf{Motion Stability.} Responding is not enough if motion stutters, stalls, or decays to a halt. Three factors penalize these---rarely moving, moving in fits and starts, and slowing to a stop---and multiply, so any one pulls the score down:
\begin{equation}
\mathrm{MS} = \rho \cdot e^{-h/h_0} \cdot \min(1, \delta/\delta_0),
\label{eq:stability}
\end{equation}
with $g$ the $70$th percentile of segment speed, $\rho$ the time fraction above $0.4g$, $h$ the rate of stalls dipping below $0.4g$ and recovering, and $\delta$ the ratio of terminal to initial speed.

\subsubsection{World Memory}

A world repainted between frames, one drifting until its geometry contradicts itself, and one rewritten after the camera turns away are three failures with different causes and fixes. Averaging them hides all three: adjacent-frame consistency is near-saturated, long-range geometry not. Local and Revisit Memory compare appearance with DINOv2~\cite{oquab2024dinov2} patch-token means $f(\cdot)$ under $d(I,I')\!=\!1\!-\!\cos(f(I),f(I'))$, avoiding 3D reconstruction, which degrades first on the content being detected; only Global Memory reconstructs, as geometry demands.

\noindent\textbf{Local Memory.} A world that flickers, mutates, or cuts outright between adjacent moments has not held together even briefly. From the hard-cut rate $r_{\text{cut}}$ of TransNetV2~\cite{soucek2024transnet} and the fraction $r_{\text{mut}}$ of adjacent pairs exceeding an absolute threshold, $\mathrm{LM}=(1-r_{\text{cut}})(1-r_{\text{mut}})$. Frozen videos are excluded and penalized by Action Dynamics instead, so failing to move earns no credit.

\noindent\textbf{Global Memory.} Every neighboring pair can look plausible while drift accumulates until the geometry contradicts itself. The feed-forward VGGT-$\Omega$~\cite{wang2026vggt} returns poses, depth, and world points in one pass; we map the median cross-frame geometric error $\tilde e$ to $\mathrm{GM}=(1+\tilde e/\tau_g)^{-1}$, bit-identical across runs, where a SLAM back-end would both vary between runs and let its robust kernel absorb the contradictions we are measuring.

\noindent\textbf{Revisit Memory.} Smooth frames and consistent geometry do not mean the world persists: a model can turn away and generate a different street on the way back. Our round-trip trajectories trigger this by design. The $K$ outbound--return pairs are matched at equal accumulated flow, not equal time, since a model that leaves quickly and returns slowly would be compared at mismatched poses. With $\bar{d}=\frac{1}{K}\sum_k d(I^{\text{out}}_k, I^{\text{ret}}_k)$ the mean paired distance, $\mathrm{RM}=(1+\bar{d}/\tau_c)^{-1}$. Pairs are spaced uniformly in arc length, and segments failing direction, response, or gain-symmetry gates are skipped.

\subsubsection{Visual Quality}

Frame quality is what existing benchmarks already measure well; we include it to test whether action dynamics and memory reduce to appearance. Perceptual and Aesthetic Quality come from Q-Align~\cite{wu2023q}, which reads logits over quality levels rather than sampling text and is therefore deterministic. Our experiments show the three categories are far from interchangeable.

\subsection{Image Suite}
\label{sec:sec_3.2}

\begin{figure}[htbp]
    \centering
    \includegraphics[width=0.7\linewidth]{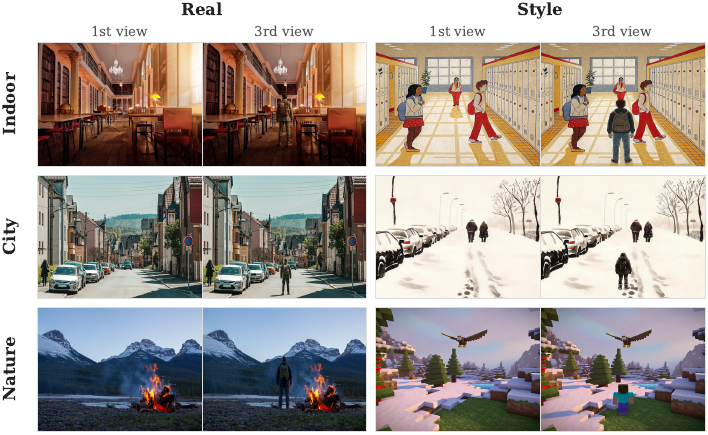}
    \caption{Overview of the Image Suite. Diverse scenes and styles are covered, shown in first and third-person views.}
    \label{fig:image}
\end{figure}

We draw 25 photorealistic and 25 stylized reference images from the 1000 per domain in the WorldScore~\cite{duan2025worldscore} pool, which already aggregates and filters many existing datasets~\cite{le2020eden, chang2017matterport3d} and so covers a broad semantic range, but redundantly: many entries repeat one scene type with minor variation. The cost of a large suite also falls on users, since comparable numbers require every evaluated model to generate the whole set, and one Hard case alone yields 60\,s of video. We therefore drop near-duplicates and keep the most representative scenes, preserving the pool's diversity at a fortieth of its size.

For each first-person reference we synthesize a matching third-person view with an image generation model~\cite{nanobana2}, giving 100 test images. The suite spans three scene categories (Nature, City, Indoor), two viewpoints, and styles from oil painting and Ukiyo-e to Minecraft.

\subsection{Action Suite}
\label{sec:sec_3.3}

\begin{figure}[htbp]
    \centering
    \includegraphics[width=0.7\linewidth]{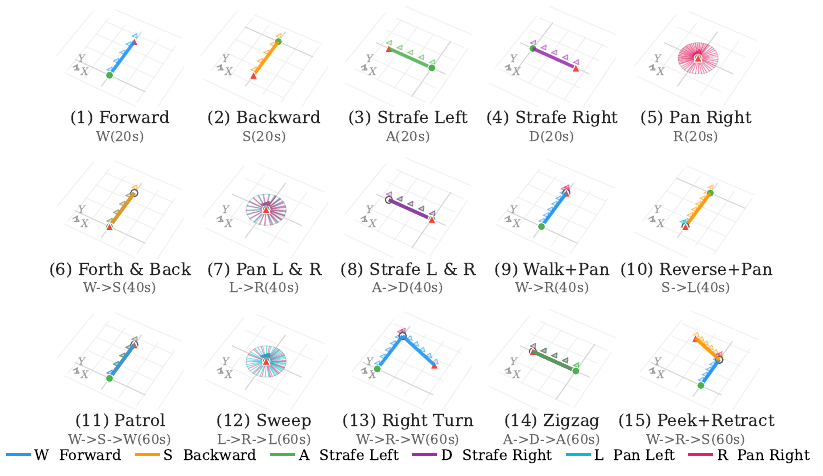}
    \caption{The 15 standardized action sequences, from elementary translations and rotations to combined trajectories.}
    \label{fig:traj_all}
\end{figure}

We define 15 action sequences (\figref{fig:traj_all}): unidirectional translations and rotations, combined translation--rotation trajectories, and cyclic round trips. All are expressed in the shared WASD+L/R vocabulary and translated per model.

The three tiers are functional rather than cosmetic: each isolates a phenomenon the previous cannot expose. \textbf{Easy}---single-segment, 20\,s---tests basic compliance and is all a model without command switching can complete. \textbf{Medium}---two-segment, 40\,s---adds exactly one switch, without which there is no transient, and provides the first round trips. \textbf{Hard}---three-segment, 60\,s, including patrol routes and $360^\circ$ rotations---runs past most models' conditioning window, where revisit and global memory fail.

\begin{figure}[htbp]
    \centering
    \includegraphics[width=0.7\linewidth]{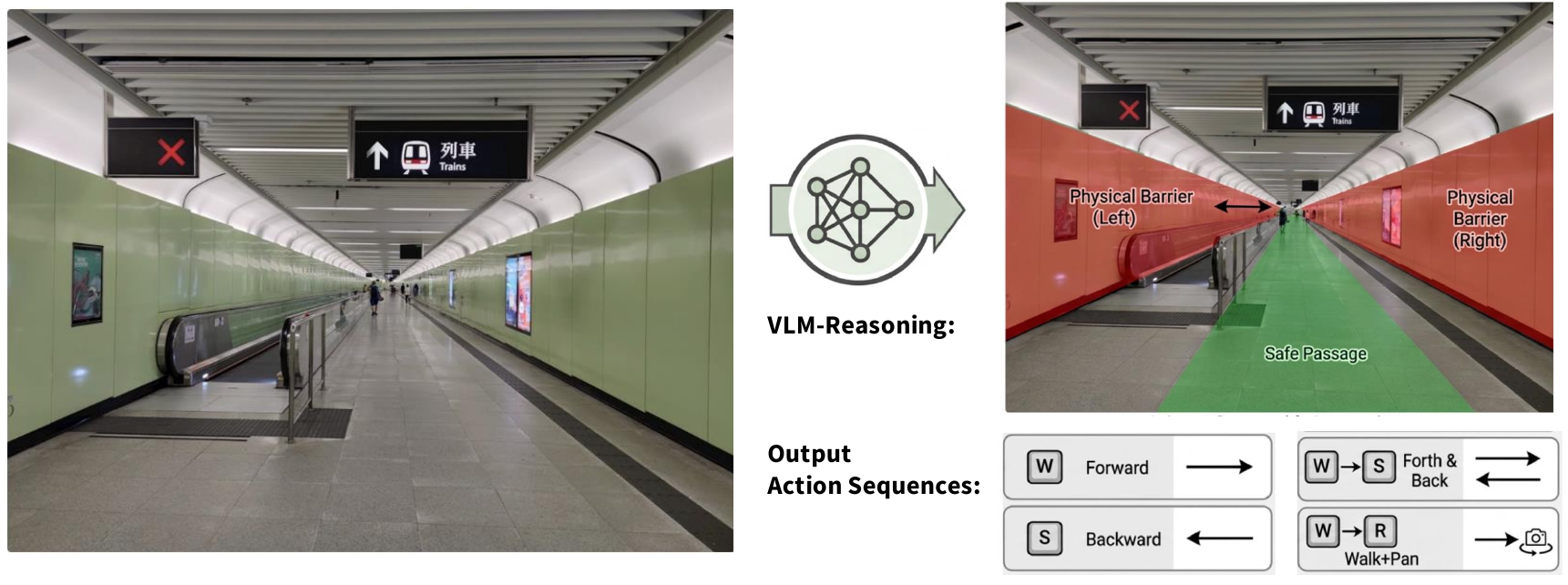}
    \caption{Context-aware action selection. A VLM analyzes the initial image to identify physical constraints and selects plausible action sequences from the predefined library.}
    \label{fig:vlm_traj}
\end{figure}

For each image a VLM~\cite{google2026gemini31} selects the five most plausible actions from the sequence library, identifying physical constraints---lateral obstacles in a corridor make prolonged strafing implausible (\figref{fig:vlm_traj})---and discarding infeasible trajectories, so models are not penalized for refusing to walk through a wall. Across the 100 test images, this yields 500 evaluation cases.

\subsection{Unified Action Interface}
\label{sec:action_interface}

The interface heterogeneity is twofold. First, models differ in \emph{what they accept}---key presses, keyboard--mouse vectors, pose strings, camera trajectories, or directional keywords in a caption---and in \emph{how they internalize} it: Plücker ray maps, PRoPE, cross-attention modules, or AdaLN-injected poses. Second, motion scales differ too: a unit step spans orders of magnitude, and yaw is per frame for some models, per action for others. The supplement lists each native format.

\method defines six primitives---forward~(W), backward~(S), strafe-left~(A), strafe-right~(D), yaw-left~(L), yaw-right~(R)---each parameterized by duration, composing all 15 trajectories. A per-model \textbf{action-mapping adapter} lowers them into the native format along three paths: models with a native keyboard interface take the primitives directly; text-driven models without a camera module take motion types and durations written into the caption; models accepting only camera control take a per-frame unified trajectory from the action primitives. Two additional details concern semantics rather than syntax. First, each model keeps its native step size and yaw rate; with units this incommensurable, calibrating to a common scale would only substitute our assumptions for the model's own. Second, most camera-conditioned models fix or estimate intrinsics internally; for the few requiring them as input, we estimate uniformly with $\pi^3$~\cite{wang2025pi}.


\begin{table}[t]
  \centering
  \caption{Quantitative comparison on the \textbf{First-Person Real and Stylized} splits (125 videos per model per split). All metrics are normalized to $[0,100]$, higher is better. Best results are in \textbf{bold} and second best are \underline{underlined}.}
  \label{tab:first-person}
  \setlength{\tabcolsep}{4.5pt}
  \renewcommand{\arraystretch}{1.12}
  \resizebox{\textwidth}{!}{%
  \begin{tabular}{l cc cc cc cc c c c c c}
    \toprule
    \multirow{3}{*}{\textbf{Model}}
      & \multicolumn{8}{c}{\textbf{Action Dynamics}}
      & \multicolumn{3}{c}{\textbf{World Memory}}
      & \multicolumn{2}{c}{\textbf{Visual Quality}} \\
    \cmidrule(lr){2-9} \cmidrule(lr){10-12} \cmidrule(lr){13-14}
      & \multicolumn{2}{c}{\makecell{Direction\\Accuracy}}
      & \multicolumn{2}{c}{\makecell{Direction\\Purity}}
      & \multicolumn{2}{c}{\makecell{Motion\\Stability}}
      & \multicolumn{2}{c}{\makecell{Response\\Latency}}
      & \multirow{2}{*}{\makecell{Local\\Memory}}
      & \multirow{2}{*}{\makecell{Global\\Memory}}
      & \multirow{2}{*}{\makecell{Revisit\\Memory}}
      & \multirow{2}{*}{\makecell{Perceptual\\Quality}}
      & \multirow{2}{*}{\makecell{Aesthetic\\Quality}} \\
    \cmidrule(lr){2-3} \cmidrule(lr){4-5} \cmidrule(lr){6-7} \cmidrule(lr){8-9}
      & trans & rot & trans & rot & trans & rot & trans & rot & & & & & \\
    \midrule
    \multicolumn{14}{l}{\textit{First-Person Real}} \\[1pt]
    AlayaWorld
      & 78.67 & \textbf{92.31} & 72.55 & \textbf{90.32} & 68.05 & \textbf{98.08} & 75.48 & 96.76 & 92.53 & \textbf{75.77} & \textbf{93.43} & \underline{85.94} & \underline{58.50} \\
    DreamX-World
      & 96.91 & 76.60 & 78.18 & 70.38 & 65.41 & 27.71 & \underline{96.10} & 97.00 & 93.29 & 34.04 & 79.89 & 71.16 & 51.98 \\
    HY-GameCraft 1.0
      & 89.07 & 2.74 & 74.82 & 63.07 & 66.76 & 13.70 & 90.97 & 51.99 & 91.31 & 36.54 & 74.20 & 64.55 & 54.38 \\
    HY-World 1.5
      & 92.55 & 85.05 & 81.18 & 80.31 & 31.66 & 86.39 & 85.44 & 85.13 & 94.17 & 58.88 & 87.52 & 76.15 & 56.59 \\
    LingBot-World
      & 89.14 & \underline{87.32} & 75.43 & 83.64 & 79.14 & 94.26 & 86.56 & \textbf{99.20} & 92.26 & 36.85 & 79.30 & 75.65 & 56.49 \\
    Lyra 2.0
      & \underline{98.16} & 85.44 & \textbf{86.32} & \underline{86.18} & \textbf{87.21} & 80.35 & 90.62 & 86.10 & \underline{96.91} & \underline{65.34} & \underline{93.26} & 84.03 & 57.66 \\
    Matrix-Game 2.0
      & 95.12 & 80.66 & 80.57 & 75.97 & \underline{79.89} & 78.85 & \textbf{98.23} & 86.47 & 91.31 & 40.56 & 72.22 & 71.15 & 51.00 \\
    Matrix-Game 3.0
      & \textbf{98.37} & 78.30 & \underline{84.83} & 73.38 & 76.39 & \underline{96.74} & 91.83 & 96.39 & 91.52 & 50.60 & 80.38 & 69.62 & 52.86 \\
    SANA-WM
      & 88.74 & 83.08 & 65.51 & 78.55 & 46.79 & 83.39 & 76.83 & \underline{97.41} & 91.70 & 53.45 & 82.05 & 76.76 & 53.16 \\
    Yume 1.5
      & 51.85 & 59.47 & 75.16 & 52.53 & 55.35 & 79.45 & 40.66 & 77.11 & \textbf{99.36} & 55.25 & 80.96 & \textbf{87.24} & \textbf{64.72} \\
    \midrule
    \multicolumn{14}{l}{\textit{First-Person Stylized}} \\[1pt]
    AlayaWorld
      & 71.89 & 81.17 & 75.78 & \underline{79.48} & 67.13 & 86.29 & 55.55 & 86.69 & 87.65 & \textbf{67.82} & \textbf{89.94} & \textbf{84.01} & \textbf{63.83} \\
    DreamX-World
      & \underline{96.57} & 74.62 & 77.65 & 69.56 & 57.20 & 36.43 & \underline{94.03} & \underline{96.02} & \underline{90.87} & 30.71 & 76.89 & 66.38 & 50.37 \\
    HY-GameCraft 1.0
      & 91.77 & 2.83 & 72.63 & 55.46 & 49.24 & 15.85 & 89.38 & 51.97 & 84.92 & 29.94 & 69.52 & 59.34 & 50.07 \\
    HY-World 1.5
      & 87.91 & \textbf{85.09} & 70.41 & 77.30 & 45.37 & 85.48 & 81.15 & 81.90 & 87.97 & \underline{52.80} & \underline{87.57} & 73.83 & 60.86 \\
    LingBot-World
      & 89.98 & \underline{83.20} & 72.31 & \textbf{79.81} & \textbf{72.22} & \underline{90.25} & 89.67 & \textbf{99.31} & 83.43 & 30.81 & 78.13 & 70.42 & 60.85 \\
    Lyra 2.0
      & 95.44 & 76.00 & \underline{80.07} & 75.56 & \underline{69.21} & 70.88 & 87.03 & 83.62 & 83.53 & 47.36 & 87.15 & 73.91 & 57.21 \\
    Matrix-Game 2.0
      & 92.17 & 77.34 & 74.10 & 71.91 & 67.09 & 78.72 & \textbf{94.58} & 89.26 & 82.89 & 38.13 & 70.03 & 69.42 & 49.40 \\
    Matrix-Game 3.0
      & \textbf{97.13} & 77.08 & \textbf{80.58} & 71.21 & 59.50 & \textbf{90.76} & 87.75 & 95.54 & 79.08 & 44.91 & 77.10 & 69.35 & 52.75 \\
    SANA-WM
      & 86.27 & 79.62 & 63.13 & 75.11 & 53.46 & 89.96 & 82.20 & 94.42 & 78.19 & 47.17 & 77.50 & 71.12 & 50.98 \\
    Yume 1.5
      & 54.24 & 51.77 & 72.92 & 44.09 & 56.79 & 70.80 & 34.58 & 80.00 & \textbf{96.36} & 43.34 & 76.96 & \underline{79.68} & \underline{61.29} \\
    \bottomrule
  \end{tabular}%
  }
\end{table}

\subsection{Evaluation Workflow}

A run proceeds in three stages. \textbf{Test Suite Selection} draws reference images by viewpoint and style, each already paired with its five VLM-selected trajectories. \textbf{Generation} lowers those trajectories through the target model's adapter and samples the videos. \textbf{Evaluation} scores them with the nine metrics. A new model joins by implementing one adapter; third-party metrics (\eg, VBench~\cite{huang2024vbench}) can also be applied to the same videos without regenerating.

\section{Experiments}

\begin{table}[t]
  \centering
  \caption{Quantitative comparison on the \textbf{Third-Person Real and Stylized} splits (125 videos per model per split); only models with native third-person support are included. Best results are in \textbf{bold} and second best are \underline{underlined}.}
  \label{tab:third-person}
  \setlength{\tabcolsep}{4.5pt}
  \renewcommand{\arraystretch}{1.12}
  \resizebox{\textwidth}{!}{%
  \begin{tabular}{l cc cc cc cc c c c c c}
    \toprule
    \multirow{3}{*}{\textbf{Model}}
      & \multicolumn{8}{c}{\textbf{Action Dynamics}}
      & \multicolumn{3}{c}{\textbf{World Memory}}
      & \multicolumn{2}{c}{\textbf{Visual Quality}} \\
    \cmidrule(lr){2-9} \cmidrule(lr){10-12} \cmidrule(lr){13-14}
      & \multicolumn{2}{c}{\makecell{Direction\\Accuracy}}
      & \multicolumn{2}{c}{\makecell{Direction\\Purity}}
      & \multicolumn{2}{c}{\makecell{Motion\\Stability}}
      & \multicolumn{2}{c}{\makecell{Response\\Latency}}
      & \multirow{2}{*}{\makecell{Local\\Memory}}
      & \multirow{2}{*}{\makecell{Global\\Memory}}
      & \multirow{2}{*}{\makecell{Revisit\\Memory}}
      & \multirow{2}{*}{\makecell{Perceptual\\Quality}}
      & \multirow{2}{*}{\makecell{Aesthetic\\Quality}} \\
    \cmidrule(lr){2-3} \cmidrule(lr){4-5} \cmidrule(lr){6-7} \cmidrule(lr){8-9}
      & trans & rot & trans & rot & trans & rot & trans & rot & & & & & \\
    \midrule
    \multicolumn{14}{l}{\textit{Third-Person Real}} \\[1pt]
    DreamX-World
      & \textbf{97.29} & 75.64 & 76.48 & 69.30 & 61.92 & 26.49 & \underline{95.98} & \underline{97.15} & \underline{93.64} & 31.16 & 79.44 & 71.60 & 52.87 \\
    HY-World 1.5
      & 90.49 & \underline{82.12} & \textbf{81.34} & \underline{77.75} & 41.92 & \textbf{87.83} & 80.69 & 84.87 & \textbf{93.92} & \textbf{55.97} & \textbf{87.55} & \underline{76.57} & \textbf{59.11} \\
    LingBot-World
      & 74.22 & 55.78 & 69.85 & 62.80 & \underline{66.76} & 64.42 & 81.15 & 74.39 & 89.89 & 47.08 & 81.68 & \textbf{76.67} & \underline{58.74} \\
    Matrix-Game 2.0
      & \underline{91.34} & 46.50 & \underline{76.49} & 39.82 & \textbf{77.59} & 75.22 & \textbf{98.11} & 87.71 & 78.76 & 31.68 & 75.64 & 67.23 & 45.78 \\
    SANA-WM
      & 88.89 & \textbf{83.10} & 64.60 & \textbf{78.33} & 46.03 & \underline{82.52} & 76.84 & \textbf{97.35} & 91.32 & \underline{52.25} & \underline{82.62} & 75.44 & 52.72 \\
    \midrule
    \multicolumn{14}{l}{\textit{Third-Person Stylized}} \\[1pt]
    DreamX-World
      & \textbf{96.39} & 75.03 & \underline{75.93} & 69.74 & 53.15 & 36.33 & \textbf{95.45} & \textbf{94.58} & \textbf{90.12} & 29.94 & \underline{77.59} & 66.22 & 50.76 \\
    HY-World 1.5
      & 93.02 & \underline{79.07} & 72.05 & \underline{71.65} & 50.15 & \underline{85.46} & 79.25 & 83.03 & \underline{89.26} & \textbf{48.04} & \textbf{85.46} & \underline{75.92} & \underline{64.05} \\
    LingBot-World
      & 69.93 & 30.90 & 65.97 & 43.49 & \underline{59.67} & 32.74 & 78.10 & 52.17 & 87.59 & \underline{47.71} & 76.87 & \textbf{77.52} & \textbf{68.03} \\
    Matrix-Game 2.0
      & \underline{93.94} & 42.42 & \textbf{76.95} & 37.49 & \textbf{74.03} & 67.46 & \underline{86.73} & 79.90 & 72.97 & 31.28 & 75.26 & 67.29 & 45.98 \\
    SANA-WM
      & 87.43 & \textbf{80.84} & 63.92 & \textbf{76.79} & 54.21 & \textbf{89.30} & 79.78 & \underline{93.91} & 75.78 & 46.17 & 76.33 & 70.59 & 51.63 \\
    \bottomrule
  \end{tabular}%
  }
\end{table}

\subsection{Experiment Setup}

\noindent\textbf{Evaluated Models.}
We benchmark ten interactive I2V world models spanning diverse control formats: Yume~1.5~\cite{mao2025yume}, HY-World~1.5~\cite{team2025hunyuanworld}, HY-GameCraft~1.0~\cite{li2025hunyuan}, Matrix-Game~2.0~\cite{he2025matrix} and 3.0~\cite{wang2026matrix}, LingBot-World~\cite{team2026advancing}, SANA-WM~\cite{zhu2026sana}, DreamX-World~\cite{team2026dreamx}, AlayaWorld~\cite{team2026alayaworld}, and Lyra~2.0~\cite{shen2026lyra}. For each we evaluate the distilled few-step variant for fast inference. Five support third-person viewpoints natively and appear in the third-person splits. We exclude closed commercial systems such as Genie~3~\cite{deepmind2025genie3}, which permit only manual web operation and cannot be driven reproducibly.

\noindent\textbf{Test Conditions.}
The Image Suite provides 50 scenes, 25 photorealistic and 25 stylized, each in paired first-/third-person views, yielding 100 images issued five action sequences each, for 500 cases. We use Nano-Banana-2~\cite{nanobana2} for third-person view synthesis and Gemini-3.1-Pro~\cite{google2026gemini31} for action selection. Every model receives identical reference images and semantically identical action sequences through its own adapter. Results are reported on four splits: First-Person Real / Stylized and Third-Person Real / Stylized.

\noindent\textbf{Compute.}
Evaluation runs on two NVIDIA H200 GPUs. Scoring one model on one split (125 videos) takes about one hour for all nine metrics, excluding video generation.

\subsection{Evaluation Results}
\label{sec:sec_4.2}
\tabref{tab:first-person} and \tabref{tab:third-person} report all four evaluation settings. Our key findings fall into four groups: what the action metrics reveal that a direction score cannot, what the three memory timescales reveal that one score cannot, what the viewpoint and style axes reveal that a photorealistic first-person suite cannot, and whether visual quality predicts any of it.

\noindent\textbf{Direction accuracy hides the differences that matter.}
On First-Person Real (\tabref{tab:first-person}), translational Direction Accuracy clusters at the top---eight of ten models exceed 88---suggesting a solved problem. The other columns disagree: Motion Stability spans 31.7--87.2 (translation) and 13.7--98.1 (rotation), translational Latency 40.7--98.2. Models indistinguishable by direction differ several-fold in how they reach it, which a single aggregate score discards.

\noindent\textbf{The fastest responders are the least stable.}
DreamX-World has the highest average Response Latency in the suite (96.1 translation, 97.0 rotation) yet the second-lowest rotational Stability (27.7), a gap of ${\sim}50$ points. The pattern recurs among fast responders: HY-World~1.5 and SANA-WM show the same signature at ${\sim}26$ and ${\sim}22$ points, and Yume~1.5 alone has stability exceeding latency.

\noindent\textbf{Command axes fail independently.}
HY-GameCraft~1.0 follows translation well (Direction Accuracy 89.1) but is unresponsive to rotation (2.7); its rotational Stability (13.7) and Latency (52.0) confirm the axis is inoperative rather than imprecise. Milder ${\sim}20$-point asymmetries appear in DreamX-World and Matrix-Game~3.0, and AlayaWorld inverts the pattern, with rotation (92.3) ahead of translation (78.7).

\noindent\textbf{Memory needs its three timescales.}
Local Memory is saturated (eight of ten models within 8 points of the best, $\sigma=2.6$) while Global Memory spreads over 34.0--75.8 ($\sigma=13.1$). An averaged memory score would follow the saturated component, concealing that DreamX-World, among the smoothest frame-to-frame (93.3), yields the least coherent global geometry (34.0). Short-horizon smoothness is nearly solved; long-range consistency is not.

\noindent\textbf{Third-person viewpoints attack rotation.}
Across the five models in both tables, the third-person setting costs $13.9$ points of rotational Direction Accuracy and $12.2$ of rotational Purity on average, against $4.0$ and $2.4$ translationally---and unevenly: Matrix-Game~2.0 loses $34.2$ and LingBot-World $31.5$, while SANA-WM and DreamX-World are unaffected. Camera control around a visible character is a distinct capability, visible only per axis and viewpoint.

\noindent\textbf{Stylization attacks memory, not action.}
From Real to Stylized, all ten models lose Local Memory ($-7.9$ mean) and all ten lose Global Memory ($-7.4$ mean, $-18.0$ worst), while Action Dynamics barely moves ($-1.5$ for translational Direction Accuracy). Command following is largely style-agnostic; keeping a stylized world self-consistent is not.

\noindent\textbf{Visual quality is not a proxy for controllability.}
Perceptual Quality correlates positively with all three memory metrics ($\rho\!=\!0.76$--$0.79$) yet negatively with translational action dynamics, most strongly with Response Latency ($\rho\!=\!-0.82$), while rotational correlations are weakly positive. Yume~1.5 is the clearest case: first on Perceptual Quality (87.2) and Local Memory (99.4), last on translational Direction Accuracy (51.9) and Latency (40.7). Whatever buys frame quality appears to cost responsiveness: ranking world models by appearance misranks them as interactive environments. 

\begin{figure}[htbp]
    \centering
    \includegraphics[width=0.7\linewidth]{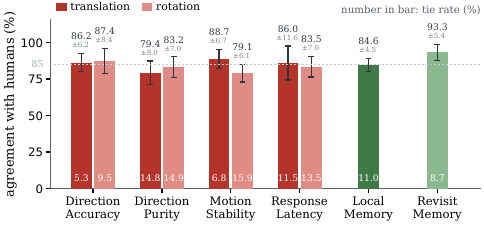}
    \caption{Metric--human agreement, per motion axis for the four action metrics. Bar height is the fraction of trials where our metric and the annotator preferred the same video, counted over trials in which a preference was expressed; the value is printed above each bar and error bars are $\pm1$ s.d.\ across the 40 annotators. The number inside each bar is the rate at which annotators could not separate the pair.}
    \label{fig:human}
\end{figure}

\subsection{Human Preference Alignment}
To establish that our new metrics are reliable, we test whether the pairs they separate by more than 10 points are the pairs humans separate the same way; Global Memory is excluded because cross-frame geometry cannot be judged by eye, and Visual Quality because it is already widely validated. Forty annotators each compared 20 blinded pairs per metric (\figref{fig:human}). Averaged over the ten dimensions, agreement is 85.1\%, from 79.1\% (rotational Stability) to 93.3\% (Revisit Memory). Ties are more frequent on rotation than translation (13.4\% vs 9.6\% over the action metrics), so rotational control is harder to judge by eye. Details are in the supplement.

\section{Conclusion and Limitations}

We presented \method, a unified benchmark for interactive I2V world models. Per-model adapters drive ten control formats with identical actions over 500 cases, and nine deterministic metrics treat action following as a step response, measuring direction, purity, latency, and stability per axis, alongside memory and visual quality. This matters: models indistinguishable by direction accuracy differ several-fold in latency, the fastest responders are often the least stable, and the best-looking model is among the least responsive.

Two dimensions are deliberately out of scope. We do not score object dynamics within the scene, which entangle with camera-induced motion; existing benchmarks~\cite{duan2025worldscore} already cover them. We also omit event-based interaction, which today is largely confined to text-to-video models and becomes measurable as interactive systems adopt it.

\bibliographystyle{abbrv}
\bibliography{references}

@String(CVPR  = {IEEE Conf. Comput. Vis. Pattern Recog.})

@String(NeurIPS = {Adv. Neural Inform. Process. Syst.})

@String(ICLR  = {Int. Conf. Learn. Represent.})

@String(AAAI  = {AAAI})

@String(CVPR  = {CVPR})

@String(NeurIPS = {NeurIPS})

@String(ICLR  = {ICLR})

@article{ho2022video,
  title={Video diffusion models},
  author={Ho, Jonathan and Salimans, Tim and Gritsenko, Alexey and Chan, William and Norouzi, Mohammad and Fleet, David J},
  journal={Advances in neural information processing systems},
  volume={35},
  pages={8633--8646},
  year={2022}
}

@article{liu2024sora,
  title={Sora: A review on background, technology, limitations, and opportunities of large vision models},
  author={Liu, Yixin and Zhang, Kai and Li, Yuan and Yan, Zhiling and Gao, Chujie and Chen, Ruoxi and Yuan, Zhengqing and Huang, Yue and Sun, Hanchi and Gao, Jianfeng and others},
  journal={arXiv preprint arXiv:2402.17177},
  year={2024}
}

@inproceedings{yangcogvideox,
  title={CogVideoX: Text-to-Video Diffusion Models with An Expert Transformer},
  author={Yang, Zhuoyi and Teng, Jiayan and Zheng, Wendi and Ding, Ming and Huang, Shiyu and Xu, Jiazheng and Yang, Yuanming and Hong, Wenyi and Zhang, Xiaohan and Feng, Guanyu and others},
  booktitle={The Thirteenth International Conference on Learning Representations},
  year      = {2025},
}

@article{kong2024hunyuanvideo,
  title={Hunyuanvideo: A systematic framework for large video generative models},
  author={Kong, Weijie and Tian, Qi and Zhang, Zijian and Min, Rox and Dai, Zuozhuo and Zhou, Jin and Xiong, Jiangfeng and Li, Xin and Wu, Bo and Zhang, Jianwei and others},
  journal={arXiv preprint arXiv:2412.03603},
  year={2024}
}

@article{wan2025wan,
  title={Wan: Open and advanced large-scale video generative models},
  author={Wan, Team and Wang, Ang and Ai, Baole and Wen, Bin and Mao, Chaojie and Xie, Chen-Wei and Chen, Di and Yu, Feiwu and Zhao, Haiming and Yang, Jianxiao and others},
  journal={arXiv preprint arXiv:2503.20314},
  year={2025}
}

@article{guo2025mineworld,
  title={Mineworld: a real-time and open-source interactive world model on minecraft},
  author={Guo, Junliang and Ye, Yang and He, Tianyu and Wu, Haoyu and Jiang, Yushu and Pearce, Tim and Bian, Jiang},
  journal={arXiv preprint arXiv:2504.08388},
  year={2025}
}

@article{oasis2024,
  author    = {Decart and Quevedo, Julian and McIntyre, Quinn and Campbell, Spruce and Chen, Xinlei and Wachen, Robert},
  title     = {Oasis: A Universe in a Transformer},
  year      = {2024},
  url       = {https://oasis-model.github.io/}
}

@article{he2025matrix,
  title={Matrix-game 2.0: An open-source real-time and streaming interactive world model},
  author={He, Xianglong and Peng, Chunli and Liu, Zexiang and Wang, Boyang and Zhang, Yifan and Cui, Qi and Kang, Fei and Jiang, Biao and An, Mengyin and Ren, Yangyang and others},
  journal={arXiv preprint arXiv:2508.13009},
  year={2025}
}

@article{team2025hunyuanworld,
  title={Hunyuanworld 1.0: Generating immersive, explorable, and interactive 3d worlds from words or pixels},
  author={Team, HunyuanWorld and Wang, Zhenwei and Liu, Yuhao and Wu, Junta and Gu, Zixiao and Wang, Haoyuan and Zuo, Xuhui and Huang, Tianyu and Li, Wenhuan and Zhang, Sheng and others},
  journal={arXiv preprint arXiv:2507.21809},
  year={2025}
}

@article{sun2025worldplay,
  title={Worldplay: Towards long-term geometric consistency for real-time interactive world modeling},
  author={Sun, Wenqiang and Zhang, Haiyu and Wang, Haoyuan and Wu, Junta and Wang, Zehan and Wang, Zhenwei and Wang, Yunhong and Zhang, Jun and Wang, Tengfei and Guo, Chunchao},
  journal={arXiv preprint arXiv:2512.14614},
  year={2025}
}

@article{li2025hunyuan,
  title={Hunyuan-gamecraft: High-dynamic interactive game video generation with hybrid history condition},
  author={Li, Jiaqi and Tang, Junshu and Xu, Zhiyong and Wu, Longhuang and Zhou, Yuan and Shao, Shuai and Yu, Tianbao and Cao, Zhiguo and Lu, Qinglin},
  journal={arXiv preprint arXiv:2506.17201},
  volume={2},
  number={3},
  pages={6},
  year={2025}
}

@article{tang2025hunyuan,
  title={Hunyuan-GameCraft-2: Instruction-following Interactive Game World Model},
  author={Tang, Junshu and Liu, Jiacheng and Li, Jiaqi and Wu, Longhuang and Yang, Haoyu and Zhao, Penghao and Gong, Siruis and Yuan, Xiang and Shao, Shuai and Lu, Qinglin},
  journal={arXiv preprint arXiv:2511.23429},
  year={2025}
}

@article{mao2025yume,
  title={Yume-1.5: A Text-Controlled Interactive World Generation Model},
  author={Mao, Xiaofeng and Li, Zhen and Li, Chuanhao and Xu, Xiaojie and Ying, Kaining and He, Tong and Pang, Jiangmiao and Qiao, Yu and Zhang, Kaipeng},
  journal={arXiv preprint arXiv:2512.22096},
  year={2025}
}

@inproceedings{meng2025towards,
  title={Towards World Simulator: Crafting Physical Commonsense-Based Benchmark for Video Generation},
  author={Meng, Fanqing and Liao, Jiaqi and Tan, Xinyu and Lu, Quanfeng and Shao, Wenqi and Zhang, Kaipeng and Cheng, Yu and Li, Dianqi and Luo, Ping},
  booktitle={International Conference on Machine Learning},
  pages={43781--43806},
  year={2025},
}

@article{li2025worldmodelbench,
  title={Worldmodelbench: Judging video generation models as world models},
  author={Li, Dacheng and Fang, Yunhao and Chen, Yukang and Yang, Shuo and Cao, Shiyi and Wong, Justin and Luo, Michael and Wang, Xiaolong and Yin, Hongxu and Gonzalez, Joseph E and others},
  journal={arXiv preprint arXiv:2502.20694},
  year={2025}
}

@inproceedings{duan2025worldscore,
  title={Worldscore: A unified evaluation benchmark for world generation},
  author={Duan, Haoyi and Yu, Hong-Xing and Chen, Sirui and Fei-Fei, Li and Wu, Jiajun},
  booktitle={Proceedings of the IEEE/CVF International Conference on Computer Vision},
  pages={27713--27724},
  year={2025}
}

@article{lu20254dworldbench,
  title={4DWorldBench: A Comprehensive Evaluation Framework for 3D/4D World Generation Models},
  author={Lu, Yiting and Luo, Wei and Tu, Peiyan and Li, Haoran and Zhu, Hanxin and Yu, Zihao and Wang, Xingrui and Chen, Xinyi and Peng, Xinge and Li, Xin and others},
  journal={arXiv preprint arXiv:2511.19836},
  year={2025}
}

@article{zhang2025matrix,
  title={Matrix-game: Interactive world foundation model},
  author={Zhang, Yifan and Peng, Chunli and Wang, Boyang and Wang, Puyi and Zhu, Qingcheng and Kang, Fei and Jiang, Biao and Gao, Zedong and Li, Eric and Liu, Yang and others},
  journal={arXiv preprint arXiv:2506.18701},
  year={2025}
}

@misc{deepmind2025genie3,
  title  = {Genie 3},
  author = {{Google DeepMind}},
  year   = {2025},
  journal= {https://deepmind.google/models/genie/}
}

@inproceedings{huang2024vbench,
  title={Vbench: Comprehensive benchmark suite for video generative models},
  author={Huang, Ziqi and He, Yinan and Yu, Jiashuo and Zhang, Fan and Si, Chenyang and Jiang, Yuming and Zhang, Yuanhan and Wu, Tianxing and Jin, Qingyang and Chanpaisit, Nattapol and others},
  booktitle={CVPR},
  pages={21807--21818},
  year={2024}
}

@article{huang2024vbench++,
  title={VBench++: Comprehensive and Versatile Benchmark Suite for Video Generative Models},
  author={Huang, Ziqi and others},
  journal={arXiv preprint arXiv:2411.13503},
  year={2024}
}

@article{mind,
  title   = {{MIND}: Benchmarking Memory Consistency and Action Control in World Models},
  author  = {Ye, Yixuan and Lu, Xuanyu and Jiang, Yuxin and Gu, Yuchao and Zhao, Rui and Liang, Qiwei and Pan, Jiachun and Zhang, Fengda and Wu, Weijia and Wang, Alex Jinpeng},
  journal = {arXiv preprint arXiv:2602.08025},
  year    = {2026}
}

@article{Droid-slam,
  title={Droid-slam: Deep visual slam for monocular, stereo, and rgb-d cameras},
  author={Teed, Zachary and Deng, Jia},
  journal={Advances in neural information processing systems},
  volume={34},
  pages={16558--16569},
  year={2021}
}

@inproceedings{zhao2025drivedreamer,
  title={Drivedreamer-2: Llm-enhanced world models for diverse driving video generation},
  author={Zhao, Guosheng and Wang, Xiaofeng and Zhu, Zheng and Chen, Xinze and Huang, Guan and Bao, Xiaoyi and Wang, Xingang},
  booktitle={Proceedings of the AAAI Conference on Artificial Intelligence},
  volume={39},
  number={10},
  pages={10412--10420},
  year={2025}
}

@article{cen2025worldvla,
  title={Worldvla: Towards autoregressive action world model},
  author={Cen, Jun and Yu, Chaohui and Yuan, Hangjie and Jiang, Yuming and Huang, Siteng and Guo, Jiayan and Li, Xin and Song, Yibing and Luo, Hao and Wang, Fan and others},
  journal={arXiv preprint arXiv:2506.21539},
  year={2025}
}

@misc{google2026gemini31,
  title  = {Gemini 3.1 Pro},
  author = {{Google DeepMind}},
  year   = {2025},
  journal    = {https://deepmind.google/models/gemini/pro/}
}

@misc{nanobana2,
  title  = {Nano Banana 2},
  author = {{Google DeepMind}},
  year   = {2026},
  journal    = {https://deepmind.google/models/gemini-image/}
}

@inproceedings{ha2018world,
  title={World Models},
  author={Ha, David and Schmidhuber, J{\"u}rgen},
  booktitle={Advances in Neural Information Processing Systems (NeurIPS)},
  volume={31},
  year={2018}
}

@inproceedings{yang2023unisim,
  title={Learning Interactive Real-World Simulators},
  author={Yang, Mengjiao and Du, Yilun and Ghasemipour, Kamyar and Tompson, Jonathan and Schuurmans, Dale and Abbeel, Pieter},
  booktitle={International Conference on Learning Representations (ICLR)},
  year={2024}
}

@article{he2024cameractrl,
  title={CameraCtrl: Enabling Camera Control for Text-to-Video Generation},
  author={He, Hao and Xu, Yinghao and Guo, Yuwei and Wetzstein, Gordon and Dai, Bo and Li, Hongsheng and Yang, Ceyuan},
  journal={arXiv preprint arXiv:2404.02101},
  year={2024}
}

@inproceedings{wang2024sea,
  title={Sea-raft: Simple, efficient, accurate raft for optical flow},
  author={Wang, Yihan and Lipson, Lahav and Deng, Jia},
  booktitle={European Conference on Computer Vision},
  pages={36--54},
  year={2024},
  organization={Springer}
}

@article{lin2025depth,
  title={Depth anything 3: Recovering the visual space from any views},
  author={Lin, Haotong and Chen, Sili and Liew, Junhao and Chen, Donny Y and Li, Zhenyu and Shi, Guang and Feng, Jiashi and Kang, Bingyi},
  journal={arXiv preprint arXiv:2511.10647},
  year={2025}
}

@article{oquab2024dinov2,
  title={Dinov2: Learning robust visual features without supervision},
  author={Oquab, Maxime and Darcet, Timoth{\'e}e and Moutakanni, Th{\'e}o and Vo, Huy and Szafraniec, Marc and Khalidov, Vasil and Fernandez, Pierre and Haziza, Daniel and Massa, Francisco and El-Nouby, Alaaeldin and others},
  journal={Transactions on Machine Learning Research Journal},
  year={2024}
}

@inproceedings{soucek2024transnet,
  title={Transnet v2: An effective deep network architecture for fast shot transition detection},
  author={Soucek, Tom{\'a}s and Lokoc, Jakub},
  booktitle={Proceedings of the 32nd ACM International Conference on Multimedia},
  pages={11218--11221},
  year={2024}
}

@article{wang2026vggt,
  title={{VGGT}-$\Omega$},
  author={Wang, Jianyuan and Chen, Minghao and Zhang, Shangzhan and Karaev, Nikita and Sch{\"o}nberger, Johannes and Labatut, Patrick and Bojanowski, Piotr and Novotny, David and Vedaldi, Andrea and Rupprecht, Christian},
  journal={arXiv preprint arXiv:2605.15195},
  year={2026}
}

@article{wu2023q,
  title={Q-align: Teaching lmms for visual scoring via discrete text-defined levels},
  author={Wu, Haoning and Zhang, Zicheng and Zhang, Weixia and Chen, Chaofeng and Liao, Liang and Li, Chunyi and Gao, Yixuan and Wang, Annan and Zhang, Erli and Sun, Wenxiu and others},
  journal={arXiv preprint arXiv:2312.17090},
  year={2023}
}

@article{zhang2026mbench,
  title={Mbench: A comprehensive benchmark on memory capability for video world models},
  author={Zhang, Shengjun and Zhang, Zhang and Huang, Simin and Tang, Zhenyu and Wang, Hanyang and Dai, Chensheng and Chen, Min and Li, Yifan and Li, Yuxin and Chen, Yingjie and others},
  journal={arXiv preprint arXiv:2606.00793},
  year={2026}
}

@inproceedings{fang2026iworld,
  title={iWorld-Bench: A Benchmark for Interactive World Models with a Unified Action Generation Framework},
  author={Fang, Jianjie and Lei, Yingshan and Wan, Qin and Wang, Ziyou and Huang, Yuchao and Xu, Yongyan and Zhao, Baining and Zhang, Weichen and Gao, Chen and Chen, Xinlei and others},
  booktitle={Forty-third International Conference on Machine Learning},
  year={2026}
}

@article{ying2026wbench,
  title={Wbench: A comprehensive multi-turn benchmark for interactive video world model evaluation},
  author={Ying, Kaining and Hu, Hengrui and Ren, Siyu and Li, Jiamu and Chen, Fengjiao and Wang, Ziwen and Cao, Xuezhi and Cai, Xunliang and Ding, Henghui},
  journal={arXiv preprint arXiv:2605.25874},
  year={2026}
}

@article{team2026alayaworld,
  title={AlayaWorld: Long-Horizon and Playable Video World Generation},
  author={Team, AlayaWorld and Zhang, Kaipeng and Li, Chuanhao and Zhan, Yifan and Ge, Yongtao and Yin, Yuanyang and Tan, Jiaming and He, Kang and Fan, Liaoyuan and Liu, Ruicong and others},
  journal={arXiv preprint arXiv:2607.06291},
  year={2026}
}

@article{team2026dreamx,
  title={DreamX-World 1.0: A General-Purpose Interactive World Model},
  author={Team, DreamX and Bai, Yancheng and Chen, Rui and Chu, Xiangxiang and Dang, Rujing and Dou, Hao and Gao, Bingjie and Gu, Qiwen and Hong, Siyu and Lei, Jiachen and others},
  journal={arXiv preprint arXiv:2606.16993},
  year={2026}
}

@article{team2026advancing,
  title={Advancing Open-source World Models},
  author={Team, Robbyant and Gao, Zelin and Wang, Qiuyu and Zeng, Yanhong and Zhu, Jiapeng and Cheng, Ka Leong and Li, Yixuan and Wang, Hanlin and Xu, Yinghao and Ma, Shuailei and others},
  journal={arXiv preprint arXiv:2601.20540},
  year={2026}
}

@article{shen2026lyra,
  title={Lyra 2.0: Explorable generative 3d worlds},
  author={Shen, Tianchang and Bahmani, Sherwin and He, Kai and Srinivasan, Sangeetha Grama and Cao, Tianshi and Ren, Jiawei and Li, Ruilong and Wang, Zian and Sharp, Nicholas and Gojcic, Zan and others},
  journal={arXiv preprint arXiv:2604.13036},
  year={2026}
}

@article{wang2026matrix,
  title={Matrix-game 3.0: Real-time and streaming interactive world model with long-horizon memory},
  author={Wang, Zile and Liu, Zexiang and Li, Jiaxing and Huang, Kaichen and Xu, Baixin and Kang, Fei and An, Mengyin and Wang, Peiyu and Jiang, Biao and Wei, Yichen and others},
  journal={arXiv preprint arXiv:2604.08995},
  year={2026}
}

@article{zhu2026sana,
  title={Sana-wm: Efficient minute-scale world modeling with hybrid linear diffusion transformer},
  author={Zhu, Haoyi and Liu, Haozhe and Zhao, Yuyang and Ye, Tian and Chen, Junsong and Yu, Jincheng and He, Tong and Han, Song and Xie, Enze},
  journal={arXiv preprint arXiv:2605.15178},
  year={2026}
}

@book{doyle2013feedback,
  title={Feedback control theory},
  author={Doyle, John C and Francis, Bruce A and Tannenbaum, Allen R},
  year={2013},
  publisher={Courier Corporation}
}

@article{huang2025vipe,
  title={Vipe: Video pose engine for 3d geometric perception},
  author={Huang, Jiahui and Zhou, Qunjie and Rabeti, Hesam and Korovko, Aleksandr and Ling, Huan and Ren, Xuanchi and Shen, Tianchang and Gao, Jun and Slepichev, Dmitry and Lin, Chen-Hsuan and others},
  journal={arXiv preprint arXiv:2508.10934},
  year={2025}
}

@article{le2020eden,
  title={Eden: Multimodal synthetic dataset of enclosed garden scenes},
  author={Le, Hoang-An and Mensink, Thomas and Das, Partha and Karaoglu, Sezer and Gevers, Theo},
  journal={arXiv preprint arXiv:2011.04389},
  year={2020}
}

@article{chang2017matterport3d,
  title={Matterport3d: Learning from rgb-d data in indoor environments},
  author={Chang, Angel and Dai, Angela and Funkhouser, Thomas and Halber, Maciej and Niessner, Matthias and Savva, Manolis and Song, Shuran and Zeng, Andy and Zhang, Yinda},
  journal={arXiv preprint arXiv:1709.06158},
  year={2017}
}

@article{wang2025pi,
  title={{$\pi^3$}: Permutation-Equivariant Visual Geometry Learning},
  author={Wang, Yifan and Zhou, Jianjun and Zhu, Haoyi and Chang, Wenzheng and Zhou, Yang and Li, Zizun and Chen, Junyi and Pang, Jiangmiao and Shen, Chunhua and He, Tong},
  journal={arXiv preprint arXiv:2507.13347},
  year={2025}
}

@inproceedings{zhu2022interiorverse,
  title={Learning-based inverse rendering of complex indoor scenes with differentiable Monte Carlo raytracing},
  author={Zhu, Jingsen and Luan, Fujun and Huo, Yuchi and Lin, Zihao and Zhong, Zhihua and Xi, Dianbing and Wang, Rui and Bao, Hujun and Zheng, Jiaxiang and Tang, Rui},
  booktitle={SIGGRAPH Asia 2022 Conference Papers},
  pages={1--8},
  year={2022}
}

@inproceedings{roberts2021hypersim,
  title={Hypersim: A photorealistic synthetic dataset for holistic indoor scene understanding},
  author={Roberts, Mike and Ramapuram, Jason and Ranjan, Anurag and Kumar, Atulit and Bautista, Miguel Angel and Paczan, Nathan and Webb, Russ and Susskind, Joshua M},
  booktitle={Proceedings of the IEEE/CVF International Conference on Computer Vision},
  pages={10912--10922},
  year={2021}
}

@inproceedings{skorokhodov2021lhq,
  title={Aligning latent and image spaces to connect the unconnectable},
  author={Skorokhodov, Ivan and Sotnikov, Grigorii and Elhoseiny, Mohamed},
  booktitle={Proceedings of the IEEE/CVF International Conference on Computer Vision},
  pages={14144--14153},
  year={2021}
}

\clearpage
\appendix
\begin{center}
  {\huge\sffamily\bfseries Supplementary Materials for WorldMark}
\end{center}
\vskip 0.6cm

This appendix supports the main paper. Sections A--C document how the
evaluation data and the per-model adapters were built; Sections D--F define
the metric constants and justify the metric design; Section~G details the
human preference alignment study; Sections H--J give additional
quantitative and qualitative results.

\section{Image Suite Construction}
\label{supp:image_suite}

\subsection{Filtering and Coverage}
\label{supp:filtering}

The Image Suite is drawn from WorldScore~\cite{duan2025worldscore}, which pools
eleven indoor and outdoor source datasets, among them
InteriorVerse, Hypersim, Matterport3D, and
LHQ~\cite{zhu2022interiorverse,roberts2021hypersim,chang2017matterport3d,skorokhodov2021lhq},
and filters them for image quality, camera perspective, mutual similarity, and
brightness, yielding 1000 photorealistic and 1000 stylized scenes over ten
subcategories per domain, five indoor and five outdoor. All are eye-level views
with no character at the frame centre, so the first-person reading is the source
condition and the lower middle is free for the third-person composite
(\secref{supp:vlm_view}).

Those filters select images that are \emph{good to look at}. Ours must also be
somewhere a camera can usefully \emph{go}, since each reference opens 20 to
60\,s of commanded navigation --- a property of scene extent that no
appearance-based filter tests. A sealed room and a featureless field both pass
every WorldScore check and fail ours: walk forward for twenty seconds in either
and nothing new appears. We therefore discard scenes too enclosed or too
uniform to reward traversal, then deduplicate by eye, keeping the 25 most
distinct per domain.

\begin{figure}[htbp]
  \centering
  \includegraphics[width=0.7\linewidth]{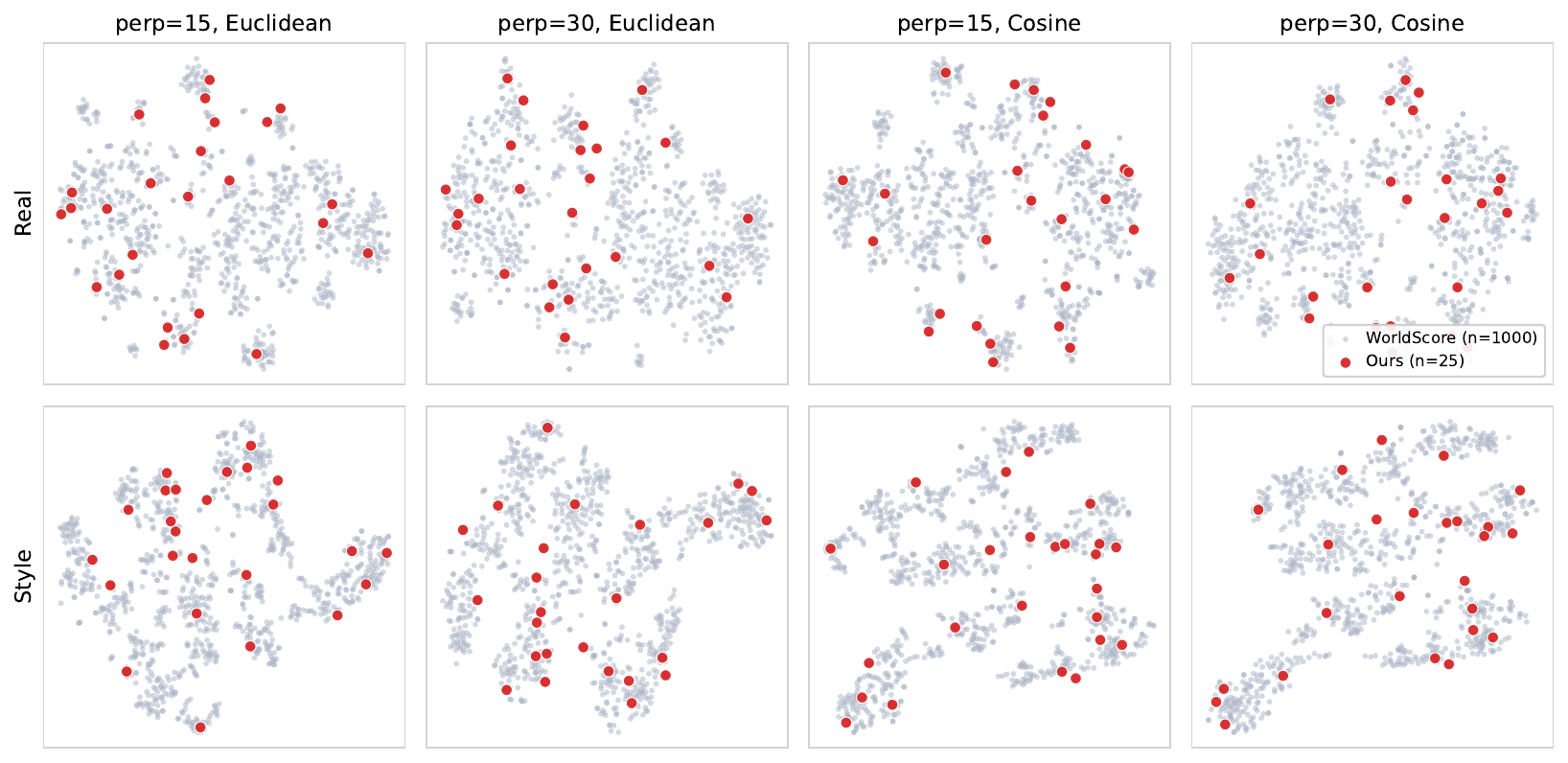}
  \caption{Coverage in CLIP feature space: the WorldScore pool in grey
           ($n = 1000$ per domain) against the retained references in red
           ($n = 25$). Rows are the two domains, columns four independent t-SNE
           projections (perplexity 15 and 30, Euclidean and cosine distance).
           The selection scatters across the occupied region in all eight
           panels.}
  \label{fig:supp_coverage}
\end{figure}

\figref{fig:supp_coverage} checks that what was discarded was repetition rather
than coverage: a fortieth of the pool reproduces its spread in all four
projections. By category the photorealistic half is 10 City, 10 Nature, 5
Indoor and the stylized half 7 City, 11 Nature, 7 Indoor --- outdoor dominates
deliberately, since an interior bounds how far a camera can travel.

\subsection{Complete Reference Set}
\label{supp:mosaic}

\begin{figure}[!p]
  \centering
  \includegraphics[width=\textwidth]{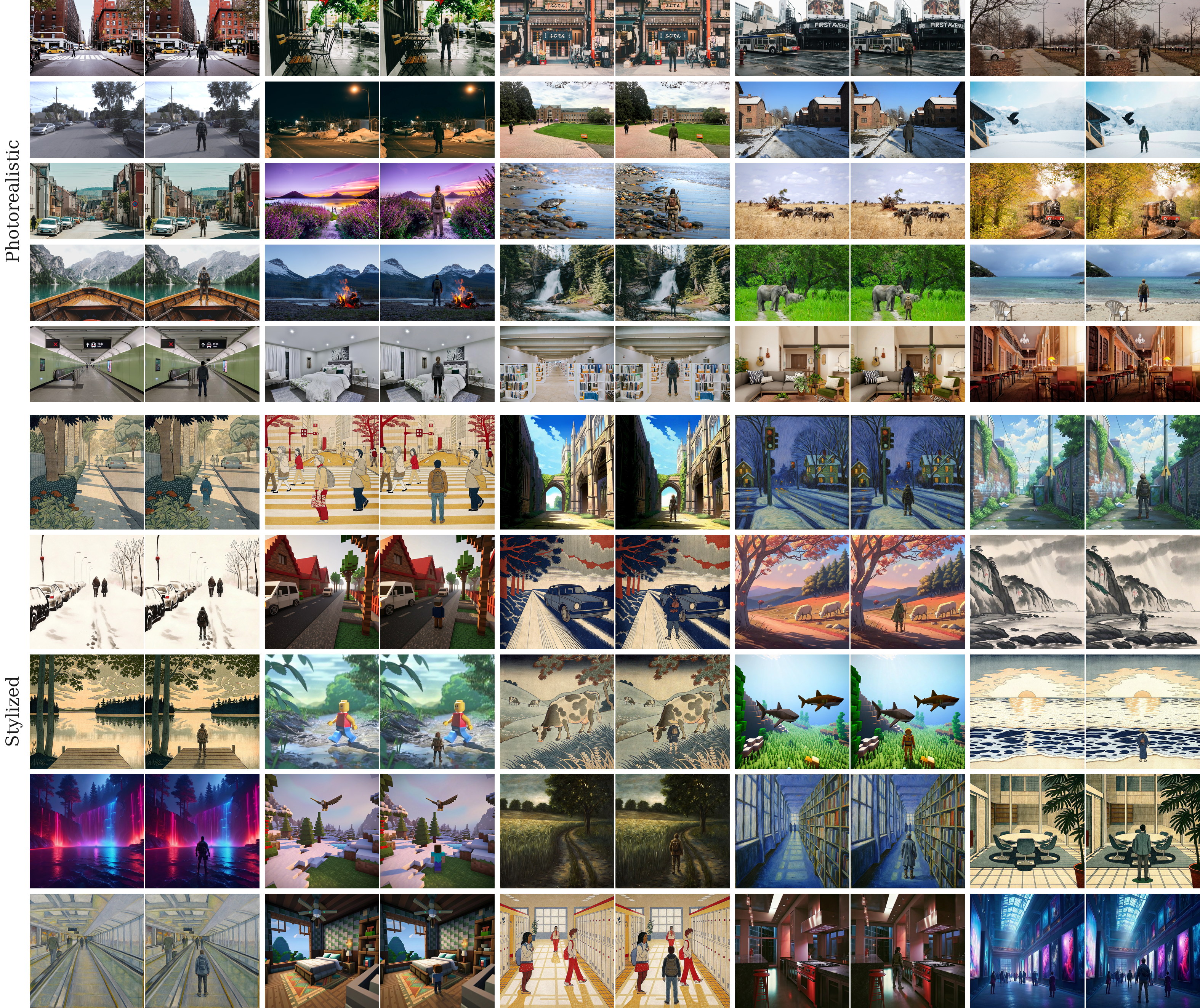}
  \caption{The complete Image Suite: all 100 test images, photorealistic above
           and stylized below. Each adjacent pair is one scene, the first-person
           reference on the left and its third-person counterpart on the right;
           the wider gutter separates scenes. Within a pair the two images agree
           everywhere except around the added figure, because the third-person
           view is the same camera with a character composited into frame rather
           than a second viewpoint.}
  \label{fig:supp_suite}
\end{figure}

\figref{fig:supp_suite} shows all 100 images as 25 scene pairs per half. The
layout makes the third-person construction plain: it is the same frame with a
character composited into the lower middle (\secref{supp:vlm_view}), not a second camera
position, so a pair agrees everywhere except around the added figure. That is
what makes the third-person splits a controlled comparison. Fifty scenes in two
viewpoints give the 100 images, and five sequences each give the 500 evaluation
cases.

The 25 stylized references span ten style families: Japanese woodblock print
(6), painting (5), voxel and Minecraft-like rendering (3), cel-shaded anime
(3), ink wash and watercolour (2), textured and vintage illustration (2), and
one each of low-poly 3D, brick-built miniature, neon-lit surrealism, and
cinematic 3D render.

\section{Prompts for Dataset Construction}
\label{supp:vlm}

\subsection{Image Suite Construction Prompts}
\label{supp:vlm_caption}

Reference images are captioned before use. Two prompts are issued, one for the
scene and one for the main character where present:

\begin{quote}\small\itshape
\textbf{Image Captioning.} Give a detailed text caption of this image. About
100 words, in one paragraph, not bullet points.

\medskip
Give a detailed text caption for the main character of this image, describing
only the appearance, in about 50 words.
\end{quote}

\subsection{Action Selection Prompt}
\label{supp:vlm_action}

For each reference image, Gemini-3.1-Pro~\cite{google2026gemini31} selects the
five most plausible action sequences from the library of 15. The prompt states
the six primitives, fixes the assumed motion scale, enumerates the library, and
constrains the output format:

\begin{quote}\small\itshape
This is a first-person view image. Suppose you are a video generation model.
Using this image as the starting point, you can perform four displacement
actions: W/A/S/D (move forward, strafe left, move backward, strafe right), and
two camera viewpoint transformation actions: L/R (turn view left, turn view
right).

\medskip
Given the 15 action sequences below, each sequence contains 1--3 actions.
Assume the movement speed is a normal human walking speed, and the camera
turning speed is 20--30 degrees per second.

\medskip
1. W (20s): Standard forward advance \\
2. S (20s): Back up and move away \\
3. A (20s): Strafe left \\
4. D (20s): Strafe right \\
5. R (20s): Look around in place \\
6. W (20s) $\rightarrow$ S (20s): Forward--backward rocking \\
7. L (20s) $\rightarrow$ R (20s): Head-shaking left and right \\
8. A (20s) $\rightarrow$ D (20s): Side-to-side hopping \\
9. W (20s) $\rightarrow$ R (20s): Walk to the end and look \\
10. S (20s) $\rightarrow$ L (20s): Reverse into a parking spot \\
11. W (20s) $\rightarrow$ S (20s) $\rightarrow$ W (20s): Patrol steps \\
12. L (20s) $\rightarrow$ R (20s) $\rightarrow$ L (20s): Left--right scanning \\
13. W (20s) $\rightarrow$ R (20s) $\rightarrow$ W (20s): Right-angle turn \\
14. A (20s) $\rightarrow$ D (20s) $\rightarrow$ A (20s): Zigzag movement \\
15. W (20s) $\rightarrow$ R (20s) $\rightarrow$ S (20s): Peek forward then retreat

\medskip
Based on this image, please provide the 5 action sequences you think are most
suitable. You need to consider environmental obstacles such as walls and
vehicles. Output only the sequence numbers directly, and do not output your
reasoning process. For example: 1, 2, 3, 4, 5
\end{quote}

\subsection{Third-Person View Synthesis Prompt}
\label{supp:vlm_view}

Third-person counterparts are synthesized with
Nano-Banana-2~\cite{nanobana2} from the same reference image:

\begin{quote}\small\itshape
Add a full-body character standing on the ground in the lower middle of the
frame, seen from behind at a distance. The character must be completely visible
from head to toe, creating a wide third-person video game perspective. Allow
modifications to the background scene elements only where they conflict with
the new character's placement. Maintain the exact original resolution.
\end{quote}

\section{Action Adapter Details}
\label{supp:adapter}

The adapter converts a benchmark sequence --- say $\mathtt{W}(20\,\mathrm{s})
\rightarrow \mathtt{R}(20\,\mathrm{s}) \rightarrow \mathtt{S}(20\,\mathrm{s})$
--- into each model's native control signal in two steps rather than one. The
sequence is first parsed into an intermediate form, a list of
(translation, yaw, duration) triples with translation in
$\{\mathrm{fwd}, \mathrm{back}, \mathrm{strafe\text{-}L}, \mathrm{strafe\text{-}R}\}$
and yaw in $\{\mathrm{left}, \mathrm{right}\}$, and only then mapped per model. The
indirection is not decoration: the same letter means different things in
different repositories, so there is no direct translation to write.
\tabref{tab:supp_formats} gives the native interface, conditioning path, and
motion scale of each model.

\begin{table}[htbp]
  \centering
  \caption{Native action interface of every evaluated model. \emph{Path} is the
           adapter route taken by the Unified Action Interface: \textsc{key} for
           models with a native keyboard interface, \textsc{txt} for
           caption-driven models, \textsc{traj} for models accepting only a
           camera trajectory. \emph{Native scale} is the model's own
           translation step and yaw increment per frame in its own units; these
           are never calibrated to a common scale, and they span a factor of
           eight in translation and five in yaw.}
  \label{tab:supp_formats}
  \small
  \resizebox{\textwidth}{!}{%
  \begin{tabular}{@{}lllllcc@{}}
    \toprule
    Model & Native action input & Conditioning mechanism & Intrinsics & Native scale (trans / yaw) & Path & 3rd \\
    \midrule
    AlayaWorld        & \texttt{camera.pt}: c2w $+$ $K$ & 3D-cache render $+$ AdaLN          & supplied, $\pi^3$ & none & \textsc{traj} & \xmark \\
    DreamX-World      & \texttt{action\_seq} $+$ frame ratios & PRoPE                        & fixed & $0.075$ / $1.5^\circ$ & \textsc{key} & \cmark \\
    HY-GameCraft 1.0  & \texttt{-{}-action-list} $+$ \texttt{-{}-action-speed-list} & Pl\"ucker $\rightarrow$ CameraNet, added to patches & fixed & $0.2$ (tunable $0$--$3$, both) & \textsc{key} & \xmark \\
    HY-World 1.5      & pose string, unit $=$ latent    & dual action representation $+$ PRoPE & fixed & $0.08$ / $3^\circ$ & \textsc{key} & \cmark \\
    LingBot-World     & \texttt{action\_string} DSL or c2w \texttt{poses.npy} & Pl\"ucker concatenated to latent & supplied, $\pi^3$ & $0.05$ / $2^\circ$ & \textsc{key} & \cmark \\
    Lyra 2.0          & \texttt{trajectory.npz} (w2c) $+$ captions & Pl\"ucker ray map, concatenated $+$ additive per block & supplied, $\pi^3$ & none & \textsc{traj} & \xmark \\
    Matrix-Game 2.0   & keyboard 4-D $+$ mouse 2-D      & ActionModule cross-attention, no camera geometry & none needed & --- & \textsc{key} & \cmark \\
    Matrix-Game 3.0   & keyboard 6-D $+$ mouse 2-D      & Pl\"ucker $+$ separate action model & derived from FOV $90^\circ$ & $12.35{\times}0.01$ / $1.5^\circ$ & \textsc{key} & \xmark \\
    SANA-WM           & \texttt{-{}-action} DSL or c2w \texttt{.npy} & Pl\"ucker raymap (additive) $+$ UCPE (PRoPE family) & self-estimated, $\pi^3$ & $0.025$ / $0.6^\circ$, smoothed & \textsc{key} & \cmark \\
    Yume 1.5          & caption text, 1 line $=2$\,s    & text prompt only, no camera module & none needed & set by text & \textsc{txt} & \xmark \\
    \bottomrule
  \end{tabular}%
  }
\end{table}

\subsection{Per-Model Adaptation}
\label{supp:adapter_permodel}

\paragraph{AlayaWorld.} The adapter writes a camera dict holding a
camera-to-world matrix per frame, the first being the identity, beside a
$3{\times}3$ intrinsic and the image and prompt files. A chunk is four latents,
$32$ frames at $24$\,fps, so a 20\,s segment is $480$ frames over fifteen
chunks.

\paragraph{DreamX-World.} Its action-speed list is not a speed but each
segment's share of the frame budget, so equal entries split the total evenly;
magnitude comes from a separate default, $0.075$ units and $1.5^\circ$ per
frame. A camera yaw is $\mathtt{l}$. A second entry point exists with
HY-GameCraft's semantics and different intrinsics, and the two cannot be mixed.

\paragraph{HY-GameCraft 1.0.} There is no duration argument: one action occupies
a fixed $33$ frames at $25$\,fps, so a segment is the same action repeated
fifteen times. Here $\mathtt{a}/\mathtt{d}$ are translation and a yaw must be
written \texttt{right\_rot}; magnitude rides on a per-action speed argument,
default $0.2$.

\paragraph{HY-World 1.5.} A pose string pairs each action with a duration in
\emph{latents}, one latent being $4$ frames at $24$\,fps, so 20\,s is $480$
frames and the string reads \texttt{w-120}. Forward motion is $0.08$ units per
frame and yaw and pitch $3^\circ$; the yaw keys are \texttt{left} and
\texttt{right}.

\paragraph{LingBot-World.} Its DSL pairs keys with a frame count, so a segment
is \texttt{w-320}; the total must be $4n{+}1$, so $960$ frames pad to $961$.
Motion is $0.05$ units and $2^\circ$ per frame with pitch clamped to $\pm
85^\circ$, and yaw sits on $\mathtt{j}/\mathtt{l}$. Intrinsics must be supplied
as a file.

\paragraph{Lyra 2.0.} Its trajectory archive carries \emph{world-to-camera}
matrices rather than camera-to-world, with a $3{\times}3$ intrinsic per frame,
and the frame count must be $1{+}80k$, so three $320$-frame segments become
$961$. Having no native scale of its own, it takes the per-frame translation and yaw we
also use for LingBot-World; the adapter integrates camera-to-world at those rates,
inverts, and tiles the intrinsic. Captions are keyed by chunk start frame; we repeat one.
Scale is read against first-frame monocular depth and can drift from the
requested path.

\paragraph{Matrix-Game 2.0.} A keyboard-and-mouse vector of the same family as
3.0's but four-dimensional rather than six, plus two mouse dimensions, at
$16$\,fps. There is no camera geometry anywhere in the model; the vector enters
through cross-attention instead.

\paragraph{Matrix-Game 3.0.} Per frame, a six-dimensional keyboard one-hot and a
two-dimensional mouse displacement. Translation is a fixed $12.35$ units per
frame, scaled by $0.01$ when the extrinsics are assembled, and the mouse
converts at $15^\circ$ per unit, so its $0.1$ step gives $1.5^\circ$. Chunks are $40$ frames except the first, which is $57$, and video
is written at $17$\,fps against a configured $16$. We drive it through a
trajectory file of one letter per line at eight chunks each.

\paragraph{SANA-WM.} $\mathtt{a}/\mathtt{d}$ are yaw and $\mathtt{j}/\mathtt{l}$
the strafe keys, transposed against every other model. Speed is not constant
either: $0.025$ units and $0.6^\circ$ per frame are targets approached by
exponential smoothing, with a $0.45$\,s rise on press and a $1$\,s coast to a
stop, so a segment accumulates slightly less than target $\times\,320$ and its
boundaries ramp rather than step.

\paragraph{Yume 1.5.} Its caption file holds one line per two-second chunk, so a
segment is ten lines, each assembled from a fixed template with independent
movement and camera slots. That separation is why rotation in place is
expressible exactly --- ``Person stands still'' beside ``Camera turns right'' ---
and the numeric tail of each line is a fixed default.

\paragraph{What this costs.} Magnitude cannot be aligned. At each model's own
defaults the same 20\,s right turn sweeps about $192^\circ$ in SANA-WM,
$480^\circ$ in Matrix-Game~3.0, $640^\circ$ in LingBot-World and $1440^\circ$ in
HY-World~1.5, while Matrix-Game~3.0 and LingBot-World normalize translation out of
their geometry branch altogether. No choice of
arguments closes that gap, which is why semantic identity is defined over action
category and temporal structure, and why Direction Accuracy and Purity read flow
through ratios in which speed cancels.

\paragraph{Segment lengths.} Where a rollout granularity does not divide 20\,s we
take the nearest integer number of units and use the \emph{same} number for
every segment, so each action in a video is given equally long. The residual is
under 1\% for HY-GameCraft; the visible exception is Matrix-Game~3.0, whose
$57$-frame opening chunk makes its first segment 19.8\,s against 18.8\,s
thereafter. We record realized frame boundaries and score on those. Segments are
neither resampled to a nominal duration nor cropped to a window common to all
models: each metric normalizes within the segment it is given, so segments
differing by a second remain comparable, whereas resampling would alter the
apparent speed.

\paragraph{Does the command arrive?} The Easy tier is the adapter's own check:
one key held for the whole segment leaves nothing to sequencing or timing, so if
the mapping is right the model moves that way. Translational Direction Accuracy
at Easy averages $88.2$ on Real with nine of ten models above $80$, and $87.2$ on
Stylized. Rotation averages $77.6$, but the shortfall is one model rather than the
mapping: excluding HY-GameCraft~1.0, whose rotation axis is inoperative rather
than imprecise, it is $84.5$. What the adapter cannot equalize is magnitude, not
direction.

\begin{table}[htbp]
  \centering
  \caption{Released checkpoint used for each model.}
  \label{tab:supp_checkpoints}
  \small
  \setlength{\tabcolsep}{3pt}
  \begin{tabular}{@{}l>{\footnotesize\raggedright\arraybackslash}p{0.60\linewidth}@{}}
    \toprule
    Model & Checkpoint \\
    \midrule
    AlayaWorld       & AlayaWorld-15B-AR-720p \\
    DreamX-World     & DreamX-World-5B-AR-720p \\
    HY-GameCraft 1.0 & Hunyuan-GameCraft-1.0-12B-720p-distilled \\
    HY-World 1.5     & HY-World1.5-AR-480P-I2V-distill \\
    LingBot-World    & LingBot-World-Fast-A14B-480P-I2V-distill \\
    Lyra 2.0         & Lyra2.0-14B-480p-I2V-DMD \\
    Matrix-Game 2.0  & MatrixGame2.0-1.8B-540p-distilled \\
    Matrix-Game 3.0  & Matrix-Game-3.0-5B-720p-distilled \\
    SANA-WM          & SANA-WM-streaming-2B-720p \\
    Yume 1.5         & Yume-5B-720p \\
    \bottomrule
  \end{tabular}
\end{table}

\tabref{tab:supp_checkpoints} gives the checkpoint used for each model. We take
the distilled few-step variant wherever one exists, since a model generates 250
cases across the two first-person splits and 250 more where third-person is
supported.

\subsection{Camera Intrinsics}
\label{supp:intrinsics}

The \emph{Intrinsics} column of \tabref{tab:supp_formats} sorts the ten models
into four regimes. Yume~1.5
and Matrix-Game~2.0 have no camera model, so intrinsics never arise. Three
carry a fixed internal calibration, and Matrix-Game~3.0 derives one exactly from
its field of view and resolution --- these are virtual cameras, with no
distortion and the principal point at the image centre. SANA-WM estimates its
own per image with $\pi^3$~\cite{wang2025pi}. The remaining three ---
AlayaWorld, LingBot-World and Lyra~2.0 --- need intrinsics from us, and we use
the same per-image $\pi^3$ estimate rather than a per-scene guess. That estimate
does not reach the scores: the Action Dynamics metrics are ratios in which focal
length cancels, so intrinsics affect what three models are conditioned on, not
what is measured on their output.

\section{Metric Thresholds and Sensitivity}
\label{supp:thresholds}

\subsection{Thresholds and Their Provenance}
\label{supp:threshold_defs}

\begin{table}[p]
  \centering
  \caption{Every constant in the \method pipeline. \textbf{S} marks a
           \emph{scoring} threshold, one that enters a reported number and could
           in principle change a ranking; lower-case \textsf{s} marks a
           \emph{structural} setting that only fixes how data is prepared. Of the
           46 entries, 17 are scoring and 27 structural; Direction Accuracy and
           Direction Purity carry no scalar threshold at all. Every value is
           fixed once and reused across all models, splits, and difficulty tiers;
           provenance for each is in \tabref{tab:supp_provenance}.}
  \label{tab:supp_thresholds}
  \footnotesize
  \setlength{\tabcolsep}{4pt}
  \begin{tabular}{@{}llllc@{}}
    \toprule
    Metric & Threshold & Value & Role & Type \\
    \midrule
    \multicolumn{5}{@{}l}{\textit{Action Dynamics}} \\[1pt]
    (shared) & flow frame stride & 2 & frames apart for SEA-RAFT flow & s \\
    (shared) & frame resize & $432{\times}640$ & input resolution & s \\
    (shared) & segment edge trim & pre 1.0s / post 3.0s & ignore transient at segment ends & s \\
    (shared) & parallax sampling & 1 s stride & frames for depth-parallax fit & s \\
    (shared) & inverse-depth span & 5-95 pct & $\Delta$($1/Z$) in $p_\mathrm{tr}$ split & s \\
    (shared) & ($1/Z$,u) point clip & 1-99 pct & outlier removal in $a,b$ fit & s \\
    Direction Accuracy & (no scalar threshold) & - & DA = (v/|v|)*w, w from $p_\mathrm{tr}$ least-squares & --- \\
    Direction Purity & (no scalar threshold) & - & DP = $E_\mathrm{cmd}$ / ($E_\mathrm{lat}$+$E_\mathrm{yaw}$+$E_\mathrm{fwd}$) & --- \\
    Response Latency & $v_\mathrm{ref}$ percentile & 75th & 'normal sustained speed' reference & \textbf{S} \\
    Response Latency & response threshold & 0.5 * $v_\mathrm{ref}$ & half sustained speed = responded & \textbf{S} \\
    Response Latency & response gate & 0.1 * $v_\mathrm{ref}$ & below $\rightarrow$ score 0 (no response) & \textbf{S} \\
    Response Latency & moving-average window & 0.5 s & smoothing of $v(t)$ & \textbf{S} \\
    Motion Stability & speed baseline $g$ & 70th pct & normal speed, robust to failures & \textbf{S} \\
    Motion Stability & sustain threshold & 0.4 * $g$ & fraction of time above $\rightarrow$ $\rho$ & \textbf{S} \\
    Motion Stability & stall hysteresis & enter 0.4g / exit 0.5g & count stalls h & \textbf{S} \\
    Motion Stability & min stall duration & 0.3 s & debounce a stall & \textbf{S} \\
    Motion Stability & stall-rate scale $h_0$ & 0.5 & in exp(-h/$h_0$) & \textbf{S} \\
    Motion Stability & decay window & first/last 3 s & terminal/initial speed ratio $\delta$ & \textbf{S} \\
    Motion Stability & decay cap & 1.2 & clip $\delta$ & \textbf{S} \\
    Motion Stability & decay target $\delta_0$ & 0.7 & in min(1, $\delta$/$\delta_0$) & \textbf{S} \\
    Motion Stability & smoothing window & 0.3 s & smoothing of $v(t)$ & \textbf{S} \\
    \midrule
    \multicolumn{5}{@{}l}{\textit{World Memory}} \\[1pt]
    Local Memory & sampling step $\delta$ & 0.5 s flow & equal-motion frame sampling & s \\
    Local Memory & frozen threshold & $<$ 5 frames & too little motion $\rightarrow$ not evaluable & s \\
    Local Memory & mutation threshold $\tau_\mathrm{mut}$ & 0.10 & DINOv2 pmean dist $>$ $\rightarrow$ a mutation & \textbf{S} \\
    Local Memory & hard-cut guard & $\pm$ 0.5 s (8 fr) & exclude pairs near a hard cut & \textbf{S} \\
    Local Memory & TransNetV2 scene thr & 0.5 & hard-cut detection & s \\
    Revisit Memory & pairs $K$ & 10 & outbound-return pairs per reversal & s \\
    Revisit Memory & min pair gap & 2.0 s & reject near-identical pairs & s \\
    Revisit Memory & arc tolerance & 30 px / 3 deg & trans / rot match tolerance & s \\
    Revisit Memory & nominal FOV & 70 deg & angle$\leftrightarrow$flow conversion (not in pairing) & s \\
    Revisit Memory & rotation magnitude gate & 15 deg & min turn to constitute a revisit & s \\
    Revisit Memory & rotation out lower bound & 5 deg & skip near-reversal frames & s \\
    Revisit Memory & direction gate & 0.3 & skip if command not followed & s \\
    Revisit Memory & gain-symmetry gate & 0.5 & skip if out/return speed asymmetric & s \\
    Revisit Memory & coverage gate & 0.3 & mark partial coverage & s \\
    Revisit Memory & return-start & 20\% steady, win 3 & onset of the return leg & s \\
    Revisit Memory & score scale $\tau_c$ & 1.0 & RM = 1/(1 + dbar/$\tau_c$) & \textbf{S} \\
    Global Memory & sampling stride & 1.0 s & frames for VGGT-Omega & s \\
    Global Memory & max frames & 64 & cap per video & s \\
    Global Memory & grid step & 16 px & point sampling grid & s \\
    Global Memory & occlusion band & 0.7-1.4 & depth-ratio occlusion test & s \\
    Global Memory & confidence drop & 30\% (CONF\_Q) & drop lowest-confidence points & s \\
    Global Memory & score scale $\tau_g$ & 0.05 & GM = 1/(1 + $\bar e$/$\tau_g$) & \textbf{S} \\
    \midrule
    \multicolumn{5}{@{}l}{\textit{Visual Quality}} \\[1pt]
    Perceptual/Aesthetic & frame sampling & 1 fps & Q-Align video mode & s \\
    Perceptual/Aesthetic & rating levels & 5 (excellent..bad) & logit-weighted 5..1 & s \\
    \midrule
    \multicolumn{5}{@{}l}{\textit{Reporting}} \\[1pt]
    (all) & normalization & [0,100] & DA/DP/MS $\times 100$, RL 100(1-tau/L), QA 100(x-1)/4 & s \\
    \bottomrule
  \end{tabular}
\end{table}

\begin{table}[t]
  \centering
  \caption{Provenance of every constant. \textsc{phys} = fixed by a physical,
           perceptual, or geometric argument; \textsc{est} = inherited from a
           backbone tool's documented configuration; \textsc{rob} = a standard
           robust-statistic convention; \textsc{norm} = a reporting convention;
           \textbf{\textsc{cal}} = calibrated from the distribution of the
           underlying measured quantity. Only the two \textsc{cal} rows read
           evaluation data at all, and both read a scalar physical quantity
           rather than model identity or rank.}
  \label{tab:supp_provenance}
  \small
  \setlength{\tabcolsep}{5pt}
  \renewcommand{\arraystretch}{1.15}
  \begin{tabular}{@{}p{0.30\textwidth} c p{0.60\textwidth}@{}}
    \toprule
    Constant(s) & Cat. & Basis \\
    \midrule
    flow stride 2, parallax 1 s stride & \textsc{phys} & inter-frame baseline giving a small but visible displacement at 16 fps \\
    segment edge trim (pre 1 s / post 3 s) & \textsc{phys} & command-onset transient + motion settling time \\
    inverse-depth 5--95 pct, $(1/Z,u)$ clip 1--99 pct & \textsc{rob} & standard outlier clipping for a least-squares fit \\
    $v_\mathrm{ref}$ = P75, speed baseline g = P70 & \textsc{rob} & robust "typical sustained speed"; upper-mid percentile resists both stalls and spikes \\
    response factor 0.5$\cdot$$v_\mathrm{ref}$ & \textsc{phys} & half of sustained speed = "the motion has begun" \\
    response gate 0.1$\cdot$$v_\mathrm{ref}$ & \textsc{phys} & essentially stationary = no response \\
    sustain 0.4$\cdot$g, hysteresis 0.4g/0.5g & \textsc{phys} & fraction-of-normal-speed cutoff; the 0.4/0.5 gap is a Schmitt-trigger margin against chatter \\
    min stall 0.3 s, smoothing 0.3 s, latency window 0.5 s & \textsc{phys} & perceptual motion-integration timescale (~0.3--0.5 s) \\
    decay window 3 s, decay cap 1.2, $h_0$ 0.5, $\delta_0$ 0.7 & \textsc{phys} & curve-shaping constants (round design values), independent of any model \\
    Local sampling $\delta$ 0.5 s, frozen < 5 frames & \textsc{phys} & equal-motion resampling step; minimum-samples guard \\
    hard-cut guard $\pm$0.5 s & \textsc{phys} & a scene cut's temporal influence window \\
    Revisit: FOV 70$^\circ$, rot gate 15$^\circ$, rot lower 5$^\circ$, arc tol 30 px / 3$^\circ$, min gap 2 s, K=10, gates 0.3/0.5/0.3, return-start 20\%/win 3 & \textsc{phys} & geometric matching tolerances and protocol design (camera FOV, degrees of turn, pixels of arc) \\
    Global: grid 16 px, occlusion band 0.7--1.4 & \textsc{phys} & point-sampling density; depth-ratio occlusion geometry \\
    frame resize 432$\times$640 & \textsc{est} & SEA-RAFT input resolution \\
    TransNetV2 scene thr 0.5 & \textsc{est} & TransNetV2's documented operating point \\
    Global: stride 1 s, max 64 frames, confidence drop 30\% & \textsc{est} & VGGT-Omega feed-forward reconstruction configuration \\
    Visual Quality: 1 fps, 5 rating levels & \textsc{est} & Q-Align video-mode convention \\
    mutation threshold $\tau_\mathrm{mut}$ = 0.10 & \textbf{\textsc{cal}} & the 95th percentile of adjacent-frame DINOv2 patch-mean distances (0.10 $\approx$ P95 = 0.101); "mutations" are by definition the rare top-5\% abrupt transitions (\figref{fig:supp_calibration}a). Per-model P95 $\in$ [0.05, 0.13] straddles it, so the cutoff is not one model's artefact \\
    Global scale $\tau_g$ = 0.05 & \textbf{\textsc{cal}} & score scale $\approx$ the mean global reprojection residual (mean $e$ = 0.055), so a typical scene maps to GM $\approx$ 0.5 (median$\rightarrow$0.58, mean$\rightarrow$0.47; \figref{fig:supp_calibration}b); monotone $\rightarrow$ cannot reorder any column \\
    Revisit scale $\tau_c$ = 1.0 & \textsc{phys} & round unit scale for the revisit residual --- all residuals $\leq$ 0.64 $\ll$ 1.0 (\figref{fig:supp_calibration}c); a fixed design constant, not distribution-fit; monotone $\rightarrow$ cannot reorder any column \\
    normalization to [0,100] & \textsc{norm} & linear rescaling of each metric's native range \\
    \bottomrule
  \end{tabular}
\end{table}

\tabref{tab:supp_thresholds} lists every constant, separating \emph{scoring} thresholds that enter a reported number from
\emph{structural} settings that only fix how data is prepared; the former are
what \secref{supp:sensitivity} sweeps. Direction Accuracy and Direction Purity have no scalar
threshold at all, being ratios of measured quantities.

\begin{figure}[htbp]
  \centering
  \includegraphics[width=\textwidth]{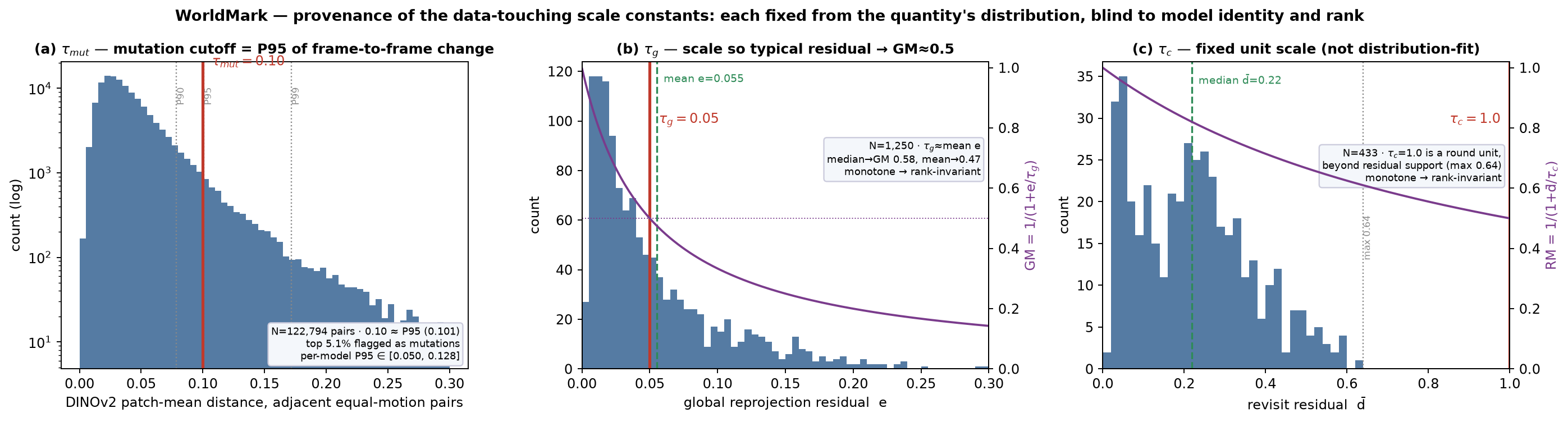}
  \caption{The constants that were checked against evaluation data. Only (a)
           and (b) are calibrated from it; (c) is a round number shown against
           its distribution for verification rather than fitted to it.
           (a) $\tau_\mathrm{mut} = 0.10$ is
           the 95th percentile of adjacent-frame DINOv2 patch-mean distance over
           122{,}794 pairs (P95 $=0.101$), so a ``mutation'' is by construction
           the rare top-5\% abrupt transition; the per-model P95 spans
           $[0.050, 0.128]$ and straddles the cutoff, so it is not one model's
           artifact. (b) $\tau_g = 0.05$ is set to the mean global reprojection
           residual ($0.055$), placing a typical scene at $\mathrm{GM} \approx
           0.5$. (c) $\tau_c = 1.0$ is a round unit scale, not a fit: every
           revisit residual lies below $0.64$, far inside it. The score maps in
           (b) and (c) are monotone, so neither constant can reorder any column.
           None of the three reads model identity or rank.}
  \label{fig:supp_calibration}
\end{figure}

\tabref{tab:supp_provenance} records where each constant came from. Most never
touch evaluation data, resting on physical or geometric arguments, backbone
defaults, or robust-statistic conventions. Exactly two are calibrated from data
(\figref{fig:supp_calibration}a, b); both read a scalar physical quantity with
model identity stripped, and one of them --- $\tau_g$ --- enters a monotone score
map that cannot reorder any column whatever value it takes, leaving
$\tau_\mathrm{mut}$ as the only genuine decision threshold. The rest are fixed
before any model is scored.

\subsection{Ranking Stability Under Perturbation}
\label{supp:sensitivity}

\begin{table}[t]
  \centering
  \caption{Ranking stability under threshold perturbation. Each of the 17
           scoring thresholds was swept over 3--5 settings straddling its
           default, one at a time with all others fixed, and the suite
           re-scored. $\rho$ is the Spearman correlation between the perturbed
           leaderboard --- the mean of the thirteen reported columns --- and the
           leaderboard under the default thresholds; $\Delta$ is the largest
           number of places any model moves. Both are worst cases over the
           sweep. Two rows move a model by more than a single place, and
           both are Motion Stability time constants; both are discussed in the
           text.}
  \label{tab:supp_sensitivity}
  \small
  \setlength{\tabcolsep}{5pt}
  \begin{tabular}{@{}llcc cc cc@{}}
    \toprule
    \multirow{2}{*}{Scoring threshold} & \multirow{2}{*}{Column} &
    \multirow{2}{*}{Default} & \multirow{2}{*}{Sweep} &
    \multicolumn{2}{c}{First-Person Real} &
    \multicolumn{2}{c}{First-Person Stylized} \\
    \cmidrule(lr){5-6} \cmidrule(lr){7-8}
    & & & & $\rho$ & $\Delta$ & $\rho$ & $\Delta$ \\
    \midrule
    Local: mutation thr $\tau_\mathrm{mut}$ & Local & 0.10 & 0.06--0.15 & 0.988 & 1 & 0.988 & 1 \\
    Local: hard-cut guard (fr) & Local & 8 & 4--16 & 1.000 & 0 & 1.000 & 0 \\
    Revisit: scale $\tau_c$ & Rev & 1.0 & 0.5--2 & 0.988 & 1 & 0.988 & 1 \\
    Global: scale $\tau_g$ & Global & 0.05 & 0.03--0.1 & 1.000 & 0 & 1.000 & 0 \\
    Latency: $v_\mathrm{ref}$ percentile & RL & 75 & 70--80 & 1.000 & 0 & 1.000 & 0 \\
    Latency: response factor & RL & 0.5 & 0.4--0.6 & 1.000 & 0 & 1.000 & 0 \\
    Latency: gate factor & RL & 0.1 & 0.05--0.15 & 1.000 & 0 & 1.000 & 0 \\
    Latency: smoothing window (s) & RL & 0.5 & 0.25--1.0 & 1.000 & 0 & 1.000 & 0 \\
    Stability: baseline g pct & MS & 70 & 60--80 & 0.988 & 1 & 0.988 & 1 \\
    Stability: sustain factor & MS & 0.4 & 0.3--0.5 & 0.988 & 1 & 0.976 & 1 \\
    Stability: stall exit hyst. ($\times g$) & MS & 0.5 & 0.45--0.6 & 1.000 & 0 & 1.000 & 0 \\
    Stability: min stall dur (s) & MS & 0.3 & 0.2--0.5 & 0.927 & 3 & 0.964 & 2 \\
    Stability: stall scale $h_0$ & MS & 0.5 & 0.4--0.6 & 1.000 & 0 & 1.000 & 0 \\
    Stability: decay window (s) & MS & 3 & 2--4 & 1.000 & 0 & 1.000 & 0 \\
    Stability: decay cap & MS & 1.2 & 1.1--1.5 & 1.000 & 0 & 1.000 & 0 \\
    Stability: decay target $\delta_0$ & MS & 0.7 & 0.6--0.8 & 0.988 & 1 & 1.000 & 0 \\
    Stability: smoothing window (s) & MS & 0.3 & 0.2--0.5 & 0.988 & 1 & 0.964 & 2 \\
    \bottomrule
  \end{tabular}
\end{table}

\tabref{tab:supp_sensitivity} reports the sweep. All 17 scoring thresholds were
swept over 3--5 settings straddling each default, one at a time, re-scoring from
cached intermediates rather than re-running any
perception model; the re-parameterised scorer is validated against the published
scores at the defaults on every run, so the sweep measures thresholds and not a
reimplementation. Fifteen hold the leaderboard to within one place on both
splits at $\rho \geq 0.976$, ten of them leaving the ordering bit-identical, and
no pair of models reverses under any setting.

Two thresholds move a model further, and both are Motion Stability time
constants left at $0.3$\,s by default. Stretching the stall debounce --- the
minimum a slowdown must persist to count as a stall --- to $0.5$\,s lifts
DreamX-World from eighth to fifth ($\rho = 0.927$); stretching the velocity
smoothing window as far moves a model two places on Stylized
($\rho = 0.964$). Neither is a fragile metric so much as a real property of the
data: DreamX-World's instability takes the form of many stalls shorter than half
a second, which a $0.5$\,s debounce erases outright. Both defaults sit inside the stable
regime with margin on either side, and that both exceptions are the same metric's
time constants is itself the finding.

\section{Metric Applicability Across Trajectories}
\label{supp:applicability}

Not every metric is defined on every trajectory, and where a metric is defined
it is not always defined on both motion axes.
\tabref{tab:supp_applicability} states this for all 15 sequences.

\begin{table}[t]
  \centering
  \caption{The unified action vocabulary and which metrics each sequence
           supports. \emph{Every segment lasts 20\,s}, so the key string fixes
           both the trajectory and its duration and the tiers are 20/40/60\,s
           (\texttt{W}/\texttt{S}/\texttt{A}/\texttt{D}: forward/backward/
           left-strafe/right-strafe; \texttt{L}/\texttt{R}: rotate left/right).
           \cmark~= applies to the whole video; T, R, and
           T+R = defined on the translational sub-metric, the
           rotational sub-metric, or both; --- = not defined for this
           trajectory type and excluded from aggregation rather than scored as
           zero. Metrics: DA Direction Accuracy, DP Direction Purity, MS Motion
           Stability, RL Response Latency, LM Local Memory, GM Global Memory,
           RM Revisit Memory, PQ Perceptual Quality, AQ Aesthetic Quality.}
  \label{tab:supp_applicability}
  \small
  \begin{tabular}{@{}lllc cccc ccc cc@{}}
    \toprule
    \multirow{2}{*}{\#} & \multirow{2}{*}{Sequence} & \multirow{2}{*}{Keys} &
    \multirow{2}{*}{Tier} &
    \multicolumn{4}{c}{Action Dynamics} &
    \multicolumn{3}{c}{World Memory} &
    \multicolumn{2}{c}{Visual Quality} \\
    \cmidrule(lr){5-8} \cmidrule(lr){9-11} \cmidrule(lr){12-13}
    & & & & DA & DP & MS & RL & LM & GM & RM & PQ & AQ \\
    \midrule
    1  & Forward       & \texttt{W}   & Easy   & T & T & T & ---        & \cmark & \cmark & ---        & \cmark & \cmark \\
    2  & Backward      & \texttt{S}   & Easy   & T & T & T & ---        & \cmark & \cmark & ---        & \cmark & \cmark \\
    3  & Strafe Left   & \texttt{A}   & Easy   & T & T & T & ---        & \cmark & \cmark & ---        & \cmark & \cmark \\
    4  & Strafe Right  & \texttt{D}   & Easy   & T & T & T & ---        & \cmark & \cmark & ---        & \cmark & \cmark \\
    5  & Pan Right     & \texttt{R}   & Easy   & R & R & R & ---        & \cmark & \cmark & ---        & \cmark & \cmark \\
    \midrule
    6  & Forth \& Back & \texttt{WS}  & Medium & T & T & T & T & \cmark & \cmark & T & \cmark & \cmark \\
    7  & Pan L \& R    & \texttt{LR}  & Medium & R & R & R & R & \cmark & \cmark & R & \cmark & \cmark \\
    8  & Strafe L \& R & \texttt{AD}  & Medium & T & T & T & T & \cmark & \cmark & T & \cmark & \cmark \\
    9  & Walk+Pan      & \texttt{WR}  & Medium & T+R & T+R & T+R & R & \cmark & \cmark & ---        & \cmark & \cmark \\
    10 & Reverse+Pan   & \texttt{SL}  & Medium & T+R & T+R & T+R & R & \cmark & \cmark & ---        & \cmark & \cmark \\
    \midrule
    11 & Patrol        & \texttt{WSW} & Hard   & T & T & T & T & \cmark & \cmark & T & \cmark & \cmark \\
    12 & Sweep         & \texttt{LRL} & Hard   & R & R & R & R & \cmark & \cmark & R & \cmark & \cmark \\
    13 & Right Turn    & \texttt{WRW} & Hard   & T+R & T+R & T+R & T+R & \cmark & \cmark & ---        & \cmark & \cmark \\
    14 & Zigzag        & \texttt{ADA} & Hard   & T & T & T & T & \cmark & \cmark & T & \cmark & \cmark \\
    15 & Peek+Retract  & \texttt{WRS} & Hard   & T+R & T+R & T+R & T+R & \cmark & \cmark & ---        & \cmark & \cmark \\
    \bottomrule
  \end{tabular}
\end{table}

Memory and quality metrics apply to the whole video and are always defined. The
four Action Dynamics metrics resolve per axis, so a sequence contributes only to
the sub-metrics its primitives exercise. Response Latency additionally needs a
command switch, which the Easy tier does not have. Revisit Memory needs the
trajectory to return along its own path, and it does so only in the six
sequences that pair a primitive with its inverse over an equal duration ---
6, 7, 8, 11, 12, and 14 --- where the return leg cancels the outbound leg
whatever native scale the model runs at. That scale invariance is the point: yaw
rates differ fivefold across the models (\secref{supp:adapter}), so a probe whose geometry
depended on the swept angle would measure something different in each of them. Sequences 9, 10, 13, and 15 interpose
a rotation between two translations, so where the second translation leads
depends on how far that model happened to turn; they are excluded from Revisit
Memory rather than scored as zero.

\section{Action Metrics Details}
\label{supp:action_metrics}

\subsection{Why We Use Optical Flow Rather Than Pose}
\label{supp:flow_vs_pose}

Response Latency times the onset of a new command and Motion Stability counts
stalls lasting about $0.3$\,s, so both need per-frame sampling and neither can be
sparsified to suppress noise. At that sampling rate the choice of signal matters:
optical flow is a measured velocity, whereas a pose stream is a reconstructed
position that must be differenced, and differencing scales noise as $1/\Delta t$.

We compared SEA-RAFT flow against two pose sources, ViPE (bundle-adjusting SLAM)
and Depth Anything 3 (feed-forward, labelled DA3 in the figure), on $30$ clips
per domain --- ten each of \texttt{LRL},
\texttt{WSW}, and \texttt{ADA}, the three sequences that reverse direction twice.
Pose direction is the first difference of the reconstructed heading for rotation
and of the camera centre for translation; flow is the raw per-frame velocity. Each
method's global axis sign was aligned once, which fixes monocular polarity and
nothing per-frame.

\begin{figure}[htbp]
  \centering
  \includegraphics[width=0.7\linewidth]{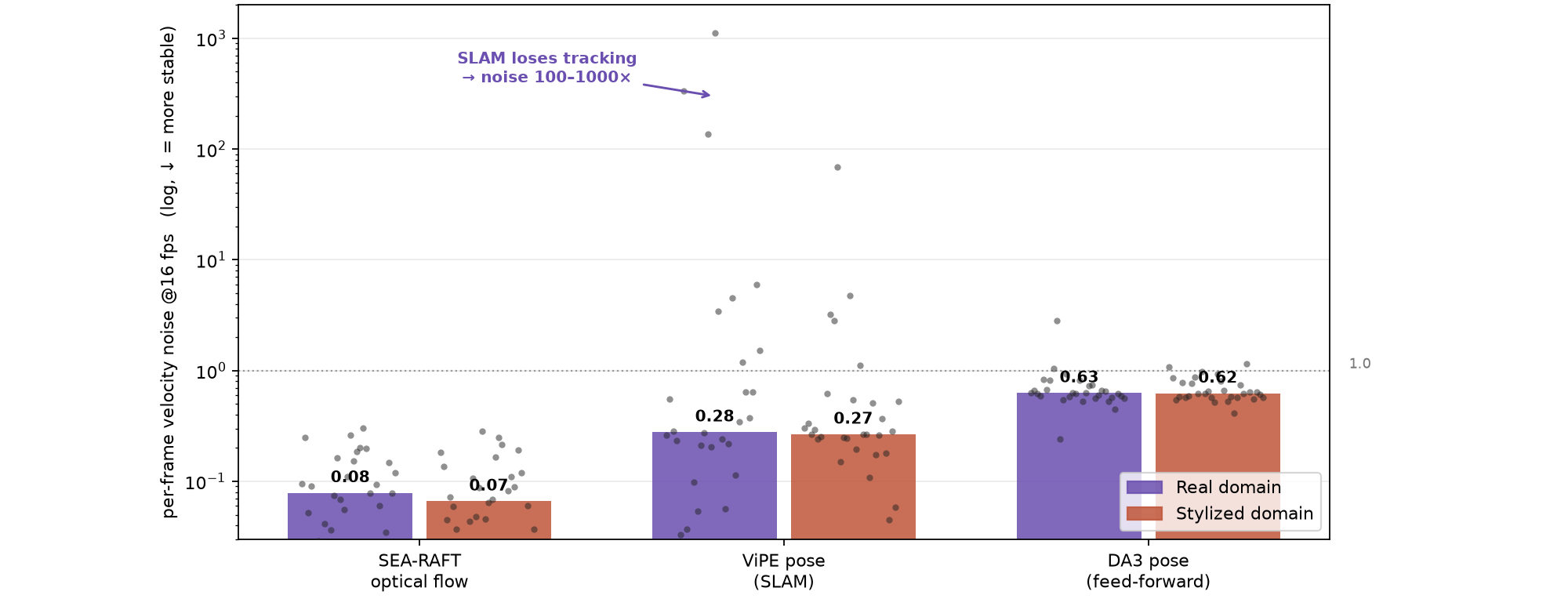}
  \caption{Per-frame velocity noise at $16$\,fps for the three candidate signals,
           on $30$ clips per domain ($10$ each of \texttt{LRL}, \texttt{WSW},
           \texttt{ADA}). Bars are medians, dots individual clips, the axis
           logarithmic. Flow is both the lowest and the tightest in each domain;
           the feed-forward estimator is uniformly noisier; the SLAM estimator is
           usually within a factor of four of flow but leaves the plot entirely on
           the clips where tracking fails.}
  \label{fig:supp_flow_vs_pose}
\end{figure}

\begin{figure}[t]
  \centering
  \includegraphics[width=0.7\linewidth]{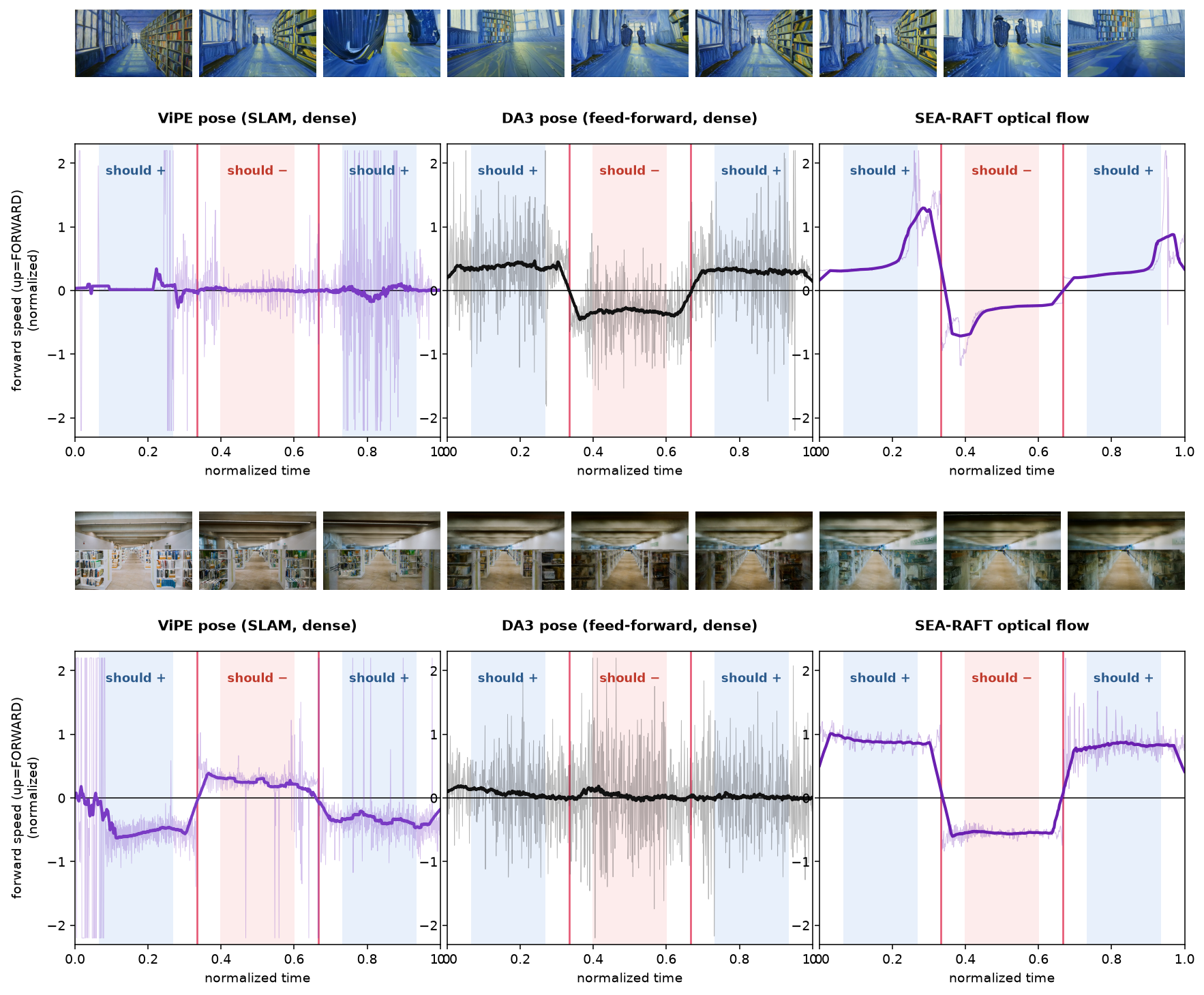}
  \caption{Two \texttt{WSW} clips, each with one pose source failing where flow
           does not. Frames above each triple; below, the per-frame directional
           velocity from SLAM, the feed-forward estimator, and flow over the
           \emph{should}~$+$~/~$-$~/~$+$ segments. \textbf{Top}, a stylized
           painterly render: SLAM's feature tracking gives way and its velocity is
           a flat trace punctuated by spikes. \textbf{Bottom}, a photorealistic
           clip whose generation washes out over the sequence: the feed-forward
           velocity is jitter about zero throughout, and SLAM's per-frame noise
           here is the $1119$ peak of \figref{fig:supp_flow_vs_pose}. Flow follows
           the reversals in both.}
  \label{fig:supp_flow_examples}
\end{figure}

Median per-frame velocity noise is $0.08$ on Real and $0.07$ on Stylized for flow,
$0.28$ and $0.27$ for SLAM, $0.63$ and $0.62$ for the feed-forward estimator
(\figref{fig:supp_flow_vs_pose}). Flow's worst clip reaches $0.3$; the
feed-forward estimator sits near eight times flow's median everywhere, and SLAM
sits near $3.5$ times it but exceeds it by two to three orders of magnitude on
roughly one clip in eight, peaking at $1119$, whenever tracking is lost.

Which source fails, and in what form, depends on the clip rather than on the motion
axis. SLAM fails in two forms: on a painterly render its feature tracking gives way
and the recovered velocity is a flat trace punctuated by spikes
(\figref{fig:supp_flow_examples}, top), while elsewhere it keeps tracking but its
per-frame noise explodes, which is where the excursions above come from. It follows
the command only where bundle adjustment fully succeeds. The feed-forward stream
fails more uniformly and for a different reason: at this sampling its per-frame pose
error dominates the true per-frame motion, so the differenced velocity is jitter
about zero --- on five of the six clips we inspected frame by frame, translation and
rotation alike, and most plainly on one whose generation washes out over the
sequence (\figref{fig:supp_flow_examples}, bottom). Flow recovers the commanded
reversal on every clip in both domains.

Those six were chosen from the sixty by ranking on low flow noise against high
pose noise, so they exhibit the failure rather than sample it; one was kept
because SLAM succeeds on it while the feed-forward stream still collapses. Depth
Anything 3 is retained in the pipeline as a depth estimator for the parallax split
of Equation~1 of the main paper, not as a source of motion.

\begin{figure}[t]
  \centering
  \includegraphics[width=0.6\linewidth]{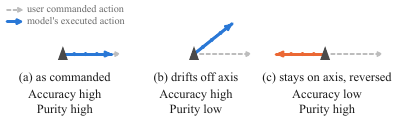}
  \caption{Why Accuracy and Purity are not interchangeable. All three cameras
           receive the same command, a strafe to the right (grey dashed); the
           solid line is the path actually taken, seen from above. In (b) the
           camera does move right, so Accuracy holds, but much of its motion is
           forward and off the commanded axis, which is what Purity penalizes.
           In (c) the motion is entirely on the commanded axis, so Purity is as
           high as in (a) --- being a ratio of magnitudes it carries no sign ---
           yet the camera runs the wrong way and Accuracy collapses. Each metric
           is blind to the failure the other catches.}
  \label{fig:supp_dadp}
\end{figure}

\subsection{What Accuracy and Purity Separate}
\label{supp:da_dp}

The two direction metrics of Equation~2 of the main paper differ in more than
emphasis. Accuracy is
signed by the commanded direction; Purity is a ratio of channel magnitudes and so
carries no sign at all (\figref{fig:supp_dadp}). A model that yaws the wrong way still keeps its motion in
the yaw channel, so it stays pure while Accuracy falls to near zero:
HY-GameCraft~1.0 reads $2.7$ Accuracy against $63.1$ Purity on rotation. Purity
alone would rate it a competent rotator; Accuracy alone would not show that the
motion it does produce is clean. Reporting both is what separates a model that
turns the wrong way from one that turns raggedly.

\section{Human Preference Alignment Details}
\label{supp:human}

\subsection{Study Design}
\label{supp:human_pairs}

Forty annotators, none involved in developing the metrics, each judged twenty
pairs for each of ten dimensions: the four action metrics per motion axis, plus
Local and Revisit Memory. Pairs were first-person cases separated by more than
$10$ points on the metric under test, a threshold fixed before any human data was
collected, shown blinded with left--right order randomized per trial. They came
from 20\,s clips holding one action, which carry the direction metrics, Motion
Stability, and Local Memory, and 40\,s round trips switching command at the
midpoint, which carry Response Latency and Revisit Memory. Revisit Memory was
shown as the matched outbound and return frames rather than as video. Clips were
scrubbable, so one annotator's $200$ judgements took roughly $40$ minutes.

Global Memory and Visual Quality were not tested: the first is cross-frame
geometric error, which is not assessable by eye, and the second uses an
established scorer rather than a metric introduced here.

\subsection{Questions and Interface}
\label{supp:study_protocol}

Six questions cover the ten dimensions. Direction Accuracy, Direction Purity,
Motion Stability, and Response Latency each split into two axes at analysis time
by grouping on the commanded action, so no question text refers to an axis. The
study was run in English; the wording below is verbatim. Quoted commands were
substituted per trial from \texttt{move forward}, \texttt{move backward},
\texttt{strafe left}, and \texttt{strafe right} for translation, and
\texttt{turn left} and \texttt{turn right} for rotation.

\begin{itemize}\itemsep2pt \parskip0pt \topsep2pt
\item \textbf{Direction Accuracy.} ``The command is \emph{move forward}. Which
      video moves in that direction?''
\item \textbf{Direction Purity}, on the same pair and command as Direction
      Accuracy. Translation: ``Same command. Which video shows only
      forward/backward movement, with less sideways drift or turning mixed in?''
      Rotation: ``Same command. Which video shows only turning, with less
      forward/backward or sideways movement mixed in?'' The two wordings differ
      because the motion that should and should not be present swaps between
      axes.
\item \textbf{Motion Stability.} ``Which video keeps moving more steadily, with
      less stalling, stuttering, or slowing to a halt?'' The three terms are the
      three factors of Equation~4 of the main paper, in order.
\item \textbf{Response Latency.} ``Halfway through, the command changes from
      \emph{move forward} to \emph{move backward}. Which video starts the new
      motion sooner?''
\item \textbf{Local Memory.} ``Which video has fewer sudden jumps or flickers
      between frames?''
\item \textbf{Revisit Memory.} ``Each pair shows the same place before the camera
      turned away and after it came back. Which pair matches more closely?''
\end{itemize}

\begin{figure}[t]
  \centering
  \includegraphics[width=0.7\linewidth]{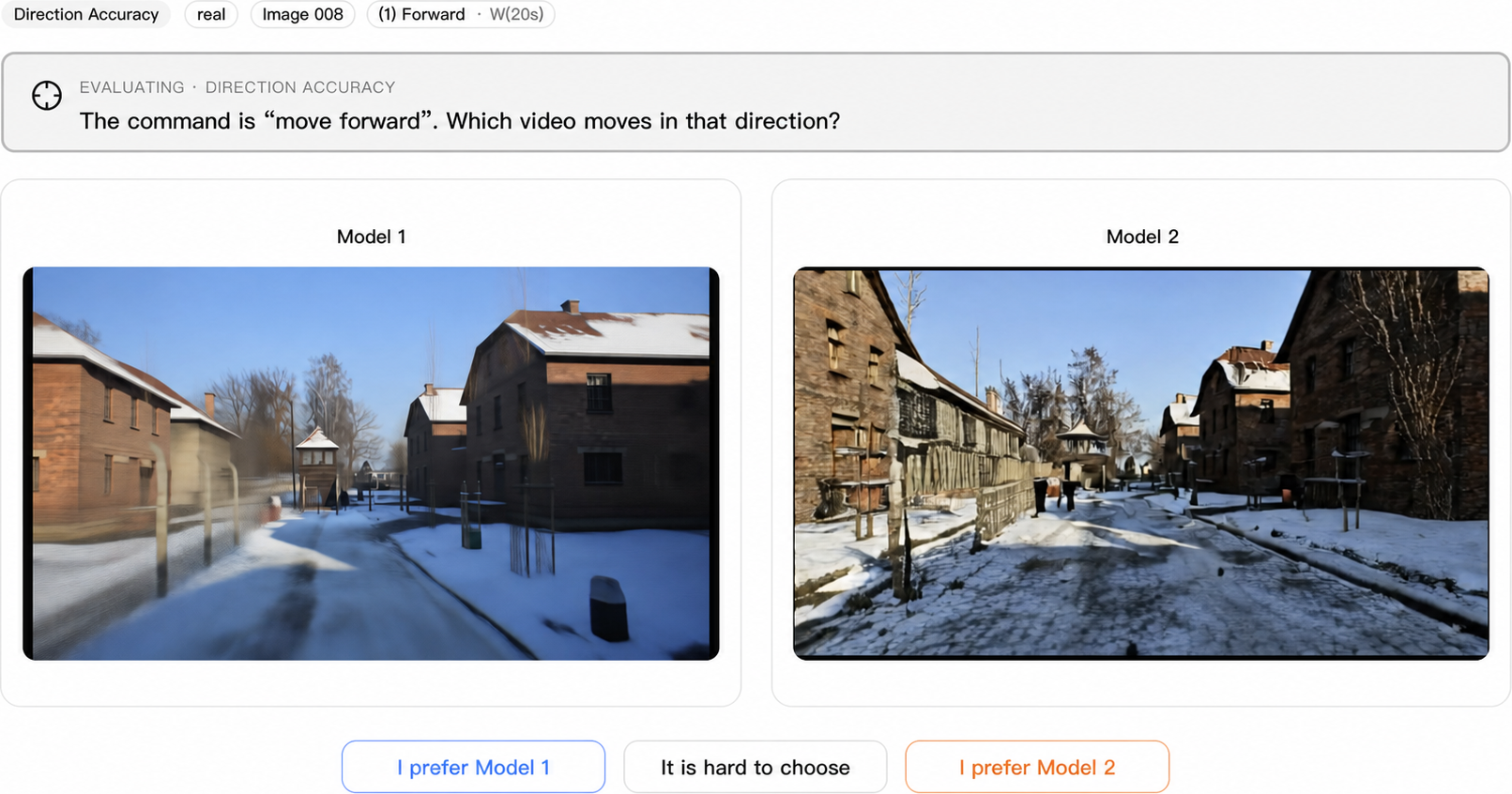}
  \caption{The annotation interface, on a Direction Accuracy trial. The header
           records the dimension, the split, the reference image, and the
           commanded sequence; the two videos are labelled only \emph{Model 1}
           and \emph{Model 2}, with left--right order randomized per trial.}
  \label{fig:supp_interface}
\end{figure}

The three response buttons read \emph{I prefer Model 1}, \emph{It is hard to
choose}, and \emph{I prefer Model 2}. \figref{fig:supp_interface} shows the
interface for a Direction Accuracy trial.

\subsection{Per-Metric Agreement}
\label{supp:human_agreement}

\begin{table}[t]
  \centering
  \caption{Metric--human agreement per dimension, regenerated from the study
           records. \emph{Agreement} is the fraction of a trial's expressed
           preferences matching the metric, computed per annotator and then
           averaged over the forty, with the standard deviation across
           annotators; $n$ is the number of trials expressing a preference, out
           of $800$. \emph{Lenient} counts ties as agreement.}
  \label{tab:supp_agreement}
  \small
  \setlength{\tabcolsep}{4pt}
  \begin{tabular}{@{}llcccc@{}}
    \toprule
    Metric & Axis & $n$ & Agreement & Tie & Lenient \\
    \midrule
    \multirow{2}{*}{Direction Accuracy} & trans & 758 & $86.2 \pm 6.2$ & 5.2 & 86.9 \\
     & rot & 724 & $87.4 \pm 8.4$ & 9.5 & 88.9 \\
    \multirow{2}{*}{Direction Purity} & trans & 682 & $79.4 \pm 8.0$ & 14.8 & 82.4 \\
     & rot & 681 & $83.2 \pm 7.0$ & 14.9 & 86.1 \\
    \multirow{2}{*}{Motion Stability} & trans & 746 & $88.7 \pm 6.7$ & 6.8 & 89.4 \\
     & rot & 673 & $79.1 \pm 6.1$ & 15.9 & 82.1 \\
    \multirow{2}{*}{Response Latency} & trans & 708 & $86.0 \pm 11.6$ & 11.5 & 88.1 \\
     & rot & 692 & $83.5 \pm 7.0$ & 13.5 & 86.1 \\
    Local Memory & --- & 712 & $84.6 \pm 4.5$ & 11.0 & 86.2 \\
    Revisit Memory & --- & 730 & $93.3 \pm 5.4$ & 8.8 & 93.9 \\
    \midrule
    \textbf{Mean} & & 710.6 & $\mathbf{85.1}$ & 11.2 & 87.0 \\
    \bottomrule
  \end{tabular}
\end{table}

Agreement, reported per dimension in \tabref{tab:supp_agreement}, is the fraction
of expressed preferences matching the metric, computed per annotator and then
averaged over the forty. It averages $85.1\%$ over the ten
dimensions, from $79.1\%$ on rotational Motion Stability to $93.3\%$ on Revisit
Memory. Outright disagreement peaks at $17.9\%$, and counting ties as agreement
gives $87.0\%$.

Regressing agreement on tie rate across the ten dimensions gives $r = -0.76$
($p = 0.011$). Averaged over its two axes Direction Purity is the weakest dimension ($79.4$ and
$83.2$) and carries the highest tie rates ($14.8$ and $14.9$). Per-annotator mean agreement spans
$81.3$--$88.7\%$ with a standard deviation of $2.0$ points, and none lies beyond
$\pm 2\sigma$.

Over the four action metrics, agreement is $85.1\%$ on translation against
$83.3\%$ on rotation, and the tie rate $9.6\%$ against $13.4\%$. In paired tests the axis
difference is significant for tie rate on Direction Accuracy ($p = 0.002$) and
Motion Stability ($p < 0.001$), and for agreement on Motion Stability alone
($p < 0.001$).

The records hold per-annotator counts of agreement, disagreement, and ties, not
which pair produced which response, so Fleiss' $\kappa$ cannot be computed from
them.

\section{Quantitative Results by Difficulty Tier}
\label{supp:per_tier}

\begin{table}[t]
  \centering
    \caption{Quantitative comparison on the \textbf{First-Person Real} split by
           difficulty tier. Normalized to $[0,100]$, higher is better; best in
           \textbf{bold} and second best \underline{underlined}, within each
           tier. Response Latency and Revisit Memory are undefined on Easy.}
  \label{tab:supp-real-difficulty}
  \setlength{\tabcolsep}{4.5pt}
  \renewcommand{\arraystretch}{1.0}
  \resizebox{\textwidth}{!}{%
  \begin{tabular}{l cc cc cc cc c c c c c}
    \toprule
    \multirow{3}{*}{\textbf{Model}}
      & \multicolumn{8}{c}{\textbf{Action Dynamics}}
      & \multicolumn{3}{c}{\textbf{World Memory}}
      & \multicolumn{2}{c}{\textbf{Visual Quality}} \\
    \cmidrule(lr){2-9} \cmidrule(lr){10-12} \cmidrule(lr){13-14}
      & \multicolumn{2}{c}{\makecell{Direction\\Accuracy}}
      & \multicolumn{2}{c}{\makecell{Direction\\Purity}}
      & \multicolumn{2}{c}{\makecell{Motion\\Stability}}
      & \multicolumn{2}{c}{\makecell{Response\\Latency}}
      & \multirow{2}{*}{\makecell{Local\\Memory}}
      & \multirow{2}{*}{\makecell{Global\\Memory}}
      & \multirow{2}{*}{\makecell{Revisit\\Memory}}
      & \multirow{2}{*}{\makecell{Perceptual\\Quality}}
      & \multirow{2}{*}{\makecell{Aesthetic\\Quality}} \\
    \cmidrule(lr){2-3} \cmidrule(lr){4-5} \cmidrule(lr){6-7} \cmidrule(lr){8-9}
      & trans & rot & trans & rot & trans & rot & trans & rot & & & & & \\
    \midrule
    \multicolumn{14}{l}{\textit{Easy (1--5)}} \\[1pt]
    AlayaWorld
      & 83.27 & \textbf{94.46} & 76.13 & \textbf{92.17} & 78.45 & \underline{99.98} & --- & --- & \underline{98.64} & \textbf{83.43} & --- & \underline{89.50} & \underline{59.88} \\
    DreamX-World
      & 94.94 & 77.59 & 77.62 & 71.67 & 61.63 & 11.01 & --- & --- & 95.65 & 49.75 & --- & 73.49 & 53.31 \\
    HY-GameCraft 1.0
      & 85.42 & 15.67 & 73.00 & 71.74 & 69.27 & 17.03 & --- & --- & 93.08 & 50.83 & --- & 69.29 & 57.01 \\
    HY-World 1.5
      & 90.68 & 89.71 & 79.67 & 84.05 & 33.99 & \textbf{100.00} & --- & --- & 95.94 & 65.82 & --- & 79.64 & 57.18 \\
    LingBot-World
      & 87.04 & 88.02 & 73.06 & 83.66 & 76.42 & 93.97 & --- & --- & 94.21 & 49.31 & --- & 80.57 & 57.41 \\
    Lyra 2.0
      & \underline{96.65} & \underline{91.11} & \textbf{86.22} & \underline{90.80} & \textbf{84.11} & 80.19 & --- & --- & 97.04 & 68.91 & --- & 84.93 & 56.81 \\
    Matrix-Game 2.0
      & 93.81 & 89.22 & 79.44 & 86.07 & \underline{78.65} & 96.62 & --- & --- & 95.28 & 61.58 & --- & 74.27 & 51.86 \\
    Matrix-Game 3.0
      & \textbf{97.46} & 84.50 & \underline{84.50} & 79.05 & 64.59 & 92.85 & --- & --- & 93.52 & 55.31 & --- & 70.11 & 51.73 \\
    SANA-WM
      & 81.21 & 81.75 & 62.06 & 78.04 & 32.30 & 67.34 & --- & --- & 91.18 & 68.00 & --- & 82.61 & 54.61 \\
    Yume 1.5
      & 71.35 & 63.86 & 72.96 & 57.21 & 57.90 & 87.81 & --- & --- & \textbf{99.80} & \underline{69.95} & --- & \textbf{90.52} & \textbf{63.95} \\
    \midrule
    \multicolumn{14}{l}{\textit{Medium (6--10)}} \\[1pt]
    AlayaWorld
      & 84.97 & \textbf{91.31} & 75.53 & \textbf{89.43} & 72.84 & \textbf{97.65} & 81.50 & 97.03 & 89.61 & \textbf{70.84} & \textbf{93.77} & 82.94 & 56.44 \\
    DreamX-World
      & 96.89 & 76.99 & 77.64 & 70.89 & 68.26 & 29.18 & 96.63 & \underline{97.40} & 90.78 & 30.50 & 79.79 & 69.09 & 50.52 \\
    HY-GameCraft 1.0
      & 90.49 & 5.56 & 75.77 & 60.63 & 67.74 & 17.44 & 93.11 & 51.63 & 90.44 & 35.62 & 72.93 & 61.47 & 53.95 \\
    HY-World 1.5
      & 94.97 & 81.65 & 84.05 & 77.37 & 29.37 & 78.97 & 84.24 & 76.50 & 94.31 & 57.08 & 87.34 & 73.77 & 56.38 \\
    LingBot-World
      & 93.02 & \underline{86.25} & 77.76 & 82.43 & 81.04 & 93.19 & 94.60 & \textbf{99.29} & 92.31 & 32.70 & 78.48 & 71.95 & 53.67 \\
    Lyra 2.0
      & \underline{98.28} & 84.22 & \textbf{86.38} & \underline{85.39} & \textbf{86.93} & 80.03 & \underline{98.28} & 87.08 & \underline{96.12} & \underline{60.90} & \underline{92.98} & \underline{83.30} & \underline{57.73} \\
    Matrix-Game 2.0
      & 93.28 & 77.20 & 82.15 & 72.11 & 74.28 & 75.26 & \textbf{98.59} & 81.31 & 92.13 & 35.89 & 71.66 & 69.17 & 50.39 \\
    Matrix-Game 3.0
      & \textbf{99.03} & 75.71 & \underline{86.37} & 70.74 & \underline{85.38} & \underline{96.98} & 95.85 & 96.51 & 92.34 & 48.42 & 79.70 & 69.31 & 53.45 \\
    SANA-WM
      & 91.96 & 84.08 & 67.59 & 79.38 & 51.92 & 88.69 & 61.99 & 96.34 & 93.28 & 49.43 & 80.99 & 73.03 & 50.98 \\
    Yume 1.5
      & 30.76 & 55.26 & 75.44 & 48.46 & 49.46 & 74.63 & 12.82 & 73.52 & \textbf{98.95} & 51.70 & 80.41 & \textbf{85.06} & \textbf{64.93} \\
    \midrule
    \multicolumn{14}{l}{\textit{Hard (11--15)}} \\[1pt]
    AlayaWorld
      & 65.55 & \textbf{92.87} & 64.55 & \textbf{90.84} & 49.29 & \underline{97.92} & 71.75 & 96.40 & 90.90 & \textbf{69.40} & \underline{92.80} & 82.99 & 59.08 \\
    DreamX-World
      & \underline{99.44} & 75.62 & 79.50 & 69.10 & 66.93 & 32.54 & \underline{95.77} & 96.48 & 92.35 & 21.88 & 80.05 & 69.44 & 51.53 \\
    HY-GameCraft 1.0
      & 92.09 & 0.00 & 76.03 & 62.98 & 62.44 & 6.91 & 89.65 & 52.46 & 88.65 & 22.21 & 76.18 & 58.82 & 48.68 \\
    HY-World 1.5
      & 92.14 & 88.02 & 79.78 & 82.98 & 31.33 & 91.44 & 86.18 & 96.18 & 89.84 & 49.07 & 87.80 & 72.15 & 55.52 \\
    LingBot-World
      & 87.36 & \underline{88.57} & 75.76 & 85.37 & 80.41 & 95.93 & 81.58 & \textbf{99.08} & 87.66 & 26.75 & 80.51 & 70.68 & \underline{59.76} \\
    Lyra 2.0
      & \textbf{99.95} & 84.85 & \textbf{86.39} & \underline{85.39} & \textbf{91.48} & 80.89 & 85.88 & 84.84 & \underline{98.15} & \underline{66.28} & \textbf{93.74} & \underline{83.23} & 59.61 \\
    Matrix-Game 2.0
      & 98.89 & 82.09 & 80.20 & 77.34 & \underline{87.94} & 76.64 & \textbf{98.01} & 93.08 & 81.25 & 25.49 & 73.08 & 67.33 & 50.05 \\
    Matrix-Game 3.0
      & 98.78 & 79.44 & \underline{83.46} & 74.82 & 81.05 & \textbf{98.02} & 89.33 & 96.23 & 85.22 & 45.06 & 81.51 & 69.03 & 54.50 \\
    SANA-WM
      & 94.60 & 82.20 & 67.50 & 77.56 & 59.35 & 82.43 & 86.02 & \underline{98.78} & 89.58 & 38.97 & 83.45 & 69.54 & 53.82 \\
    Yume 1.5
      & 51.31 & 63.71 & 77.63 & 56.45 & 58.89 & 82.94 & 57.90 & 81.70 & \textbf{99.14} & 39.86 & 81.95 & \textbf{83.38} & \textbf{66.22} \\
    \bottomrule
  \end{tabular}%
  }
\end{table}

\begin{table}[t]
  \centering
    \caption{Quantitative comparison on the \textbf{First-Person Stylized} split by
           difficulty tier. Normalized to $[0,100]$, higher is better; best in
           \textbf{bold} and second best \underline{underlined}, within each
           tier. Response Latency and Revisit Memory are undefined on Easy.}
  \label{tab:supp-style-difficulty}
  \setlength{\tabcolsep}{4.5pt}
  \renewcommand{\arraystretch}{1.0}
  \resizebox{\textwidth}{!}{%
  \begin{tabular}{l cc cc cc cc c c c c c}
    \toprule
    \multirow{3}{*}{\textbf{Model}}
      & \multicolumn{8}{c}{\textbf{Action Dynamics}}
      & \multicolumn{3}{c}{\textbf{World Memory}}
      & \multicolumn{2}{c}{\textbf{Visual Quality}} \\
    \cmidrule(lr){2-9} \cmidrule(lr){10-12} \cmidrule(lr){13-14}
      & \multicolumn{2}{c}{\makecell{Direction\\Accuracy}}
      & \multicolumn{2}{c}{\makecell{Direction\\Purity}}
      & \multicolumn{2}{c}{\makecell{Motion\\Stability}}
      & \multicolumn{2}{c}{\makecell{Response\\Latency}}
      & \multirow{2}{*}{\makecell{Local\\Memory}}
      & \multirow{2}{*}{\makecell{Global\\Memory}}
      & \multirow{2}{*}{\makecell{Revisit\\Memory}}
      & \multirow{2}{*}{\makecell{Perceptual\\Quality}}
      & \multirow{2}{*}{\makecell{Aesthetic\\Quality}} \\
    \cmidrule(lr){2-3} \cmidrule(lr){4-5} \cmidrule(lr){6-7} \cmidrule(lr){8-9}
      & trans & rot & trans & rot & trans & rot & trans & rot & & & & & \\
    \midrule
    \multicolumn{14}{l}{\textit{Easy (1--5)}} \\[1pt]
    AlayaWorld
      & 89.88 & \textbf{92.41} & 79.41 & \textbf{90.11} & \textbf{83.29} & \underline{96.56} & --- & --- & 90.84 & \textbf{79.32} & --- & \textbf{86.36} & \textbf{64.20} \\
    DreamX-World
      & 94.15 & 80.96 & 75.49 & 76.07 & 56.72 & 12.14 & --- & --- & 90.61 & 44.00 & --- & 68.64 & 51.61 \\
    HY-GameCraft 1.0
      & 90.89 & 13.90 & 73.59 & 62.05 & 52.06 & 16.09 & --- & --- & 87.52 & 48.81 & --- & 66.58 & 56.74 \\
    HY-World 1.5
      & 84.96 & \underline{86.67} & 67.80 & 76.61 & 35.39 & \textbf{99.11} & --- & --- & \underline{93.37} & 62.73 & --- & 77.79 & \underline{62.70} \\
    LingBot-World
      & 87.99 & 80.79 & 71.16 & 76.82 & \underline{72.43} & 89.46 & --- & --- & 86.35 & 40.25 & --- & 74.02 & 61.60 \\
    Lyra 2.0
      & \underline{94.98} & 85.12 & \textbf{82.48} & \underline{85.26} & 65.20 & 73.96 & --- & --- & 86.58 & 48.15 & --- & 75.87 & 57.68 \\
    Matrix-Game 2.0
      & 87.91 & 85.31 & 71.18 & 80.00 & 60.59 & 91.12 & --- & --- & 84.85 & 55.79 & --- & 70.74 & 49.01 \\
    Matrix-Game 3.0
      & \textbf{96.29} & 77.54 & \underline{81.53} & 71.46 & 50.07 & 91.44 & --- & --- & 81.57 & 54.50 & --- & 69.59 & 51.32 \\
    SANA-WM
      & 77.85 & 71.76 & 58.76 & 68.90 & 36.33 & 71.43 & --- & --- & 83.07 & \underline{66.62} & --- & 77.81 & 55.07 \\
    Yume 1.5
      & 67.57 & 60.69 & 72.76 & 55.91 & 64.24 & 80.94 & --- & --- & \textbf{97.70} & 64.44 & --- & \underline{80.91} & 60.67 \\
    \midrule
    \multicolumn{14}{l}{\textit{Medium (6--10)}} \\[1pt]
    AlayaWorld
      & 79.30 & 83.34 & \underline{79.00} & \textbf{80.91} & 66.19 & 83.98 & 70.41 & 84.83 & 88.48 & \textbf{64.02} & \textbf{90.58} & \textbf{82.59} & \textbf{63.09} \\
    DreamX-World
      & \textbf{97.41} & 74.24 & 77.92 & 68.96 & 54.86 & 35.62 & \textbf{97.61} & \underline{96.38} & \underline{91.65} & 26.01 & 75.47 & 63.93 & 48.29 \\
    HY-GameCraft 1.0
      & 96.16 & 1.86 & 76.50 & 51.99 & 51.77 & 19.23 & 89.63 & 60.98 & 82.83 & 21.82 & 68.64 & 54.39 & 45.30 \\
    HY-World 1.5
      & 94.21 & \underline{83.89} & 70.81 & 77.74 & 51.70 & 79.68 & 83.71 & 71.99 & 84.44 & \underline{47.64} & \underline{87.53} & 70.98 & 59.13 \\
    LingBot-World
      & 86.80 & \textbf{84.05} & 71.51 & \underline{80.71} & \textbf{72.74} & 87.42 & 85.72 & \textbf{99.45} & 80.39 & 26.50 & 77.55 & 68.73 & 60.20 \\
    Lyra 2.0
      & 93.45 & 70.22 & 77.66 & 70.59 & \underline{69.13} & 68.34 & 93.21 & 80.15 & 79.48 & 45.68 & 87.18 & 72.43 & 56.53 \\
    Matrix-Game 2.0
      & 93.90 & 69.13 & 74.11 & 63.36 & 66.70 & 69.42 & \underline{95.72} & 84.40 & 78.89 & 32.08 & 68.69 & 67.52 & 48.42 \\
    Matrix-Game 3.0
      & \underline{96.67} & 75.44 & \textbf{79.31} & 68.99 & 66.72 & \underline{87.50} & 91.23 & 95.06 & 76.84 & 38.82 & 76.09 & 68.93 & 53.64 \\
    SANA-WM
      & 94.42 & 82.15 & 66.34 & 77.49 & 63.98 & \textbf{92.85} & 73.38 & 92.65 & 70.73 & 40.51 & 77.14 & 66.22 & 46.91 \\
    Yume 1.5
      & 41.68 & 48.37 & 73.31 & 41.12 & 51.55 & 68.58 & 13.95 & 81.92 & \textbf{96.34} & 37.96 & 76.67 & \underline{78.58} & \underline{60.91} \\
    \midrule
    \multicolumn{14}{l}{\textit{Hard (11--15)}} \\[1pt]
    AlayaWorld
      & 30.66 & 73.05 & 64.86 & 72.77 & 41.63 & 85.51 & 42.43 & 89.18 & 79.32 & \textbf{52.54} & \textbf{88.51} & \textbf{80.36} & \textbf{64.43} \\
    DreamX-World
      & \textbf{99.33} & 72.52 & \underline{80.86} & 67.71 & 61.53 & 47.99 & 90.88 & 95.54 & \underline{89.68} & 20.82 & 79.55 & 65.34 & 51.52 \\
    HY-GameCraft 1.0
      & 86.57 & 0.00 & 65.16 & 58.02 & 40.70 & 10.53 & 89.17 & 39.84 & 82.07 & 23.31 & 71.18 & 49.38 & 41.33 \\
    HY-World 1.5
      & 83.26 & \underline{86.28} & 74.15 & 76.90 & 52.37 & 88.65 & 78.90 & 95.24 & 82.08 & 43.36 & \underline{87.64} & 68.68 & 59.44 \\
    LingBot-World
      & 98.13 & 82.92 & 75.43 & \underline{79.70} & 71.10 & \underline{94.98} & \underline{93.15} & \textbf{99.13} & 82.49 & 23.37 & 78.81 & 63.67 & 60.13 \\
    Lyra 2.0
      & 99.21 & 81.06 & 79.74 & 79.13 & \underline{75.99} & 73.50 & 81.58 & 88.30 & 83.99 & \underline{49.12} & 87.10 & 71.56 & 57.39 \\
    Matrix-Game 2.0
      & 96.62 & \textbf{86.63} & 78.97 & \textbf{81.70} & \textbf{78.53} & 87.82 & \textbf{93.57} & 95.82 & 86.81 & 25.42 & 72.38 & 69.89 & 52.83 \\
    Matrix-Game 3.0
      & \underline{99.22} & 79.42 & \textbf{80.95} & 74.54 & 64.26 & \textbf{95.49} & 84.67 & 96.20 & 77.00 & 38.78 & 79.08 & 69.58 & 54.94 \\
    SANA-WM
      & 87.94 & 79.04 & 65.52 & 74.06 & 66.04 & 93.34 & 89.98 & \underline{96.80} & 83.96 & 31.79 & 78.17 & 62.66 & 48.30 \\
    Yume 1.5
      & 51.07 & 53.24 & 72.59 & 43.66 & 52.34 & 69.92 & 52.78 & 77.41 & \textbf{92.64} & 26.43 & 77.37 & \underline{78.55} & \underline{64.03} \\
    \bottomrule
  \end{tabular}%
  }
\end{table}

\textcolor{refblue}{Tables~\ref{tab:supp-real-difficulty}
and~\ref{tab:supp-style-difficulty}} report
all thirteen columns per tier, with best and second best computed within a tier.
Response Latency and Revisit Memory are marked --- on Easy, where a single-key
action neither switches command nor returns, and are excluded from the Easy
ranking.

\paragraph{Difficulty costs long-range geometry and little else.} Averaged over
the ten models on Real, Global Memory loses 35.0\% from Easy to Hard
(62.3~$\rightarrow$~40.5) and falls for all ten (40.7\% on Stylized), while
Perceptual Quality declines by only 8.6\%, Local Memory by 5.4\%, and Aesthetic
Quality is flat at 0.9\%. This also fixes 60\,s as the end of the Action Suite: a
longer tier would separate models by the speed of their disintegration rather
than expose a new failure.

\paragraph{Off-axis contamination is not a difficulty effect.} No model reaches
90 on translational Direction Purity in \emph{any} tier on \emph{either} split,
and none reaches 85 on Stylized; the ceiling is $86.4$ and $82.5$, and Easy already tops out at $86.2$ and $82.5$,
with one key held for the whole segment. The median
Accuracy-minus-Purity gap likewise stays about three times larger
translationally at every tier ($12.7$--$14.4$ against $4.0$--$4.6$ on Real,
$16.8$--$18.4$ against $4.4$--$5.0$ on Stylized).

\paragraph{Rotation is what difficulty breaks.} Rotational Direction Accuracy
falls from Easy to Hard for eight of ten models on Real, while translational
Accuracy does not move there ($88.2~\rightarrow~88.0$).

\section{Per-Model Capability Profiles}
\label{supp:radar}

\begin{figure}[t]
  \centering
  \includegraphics[width=\textwidth]{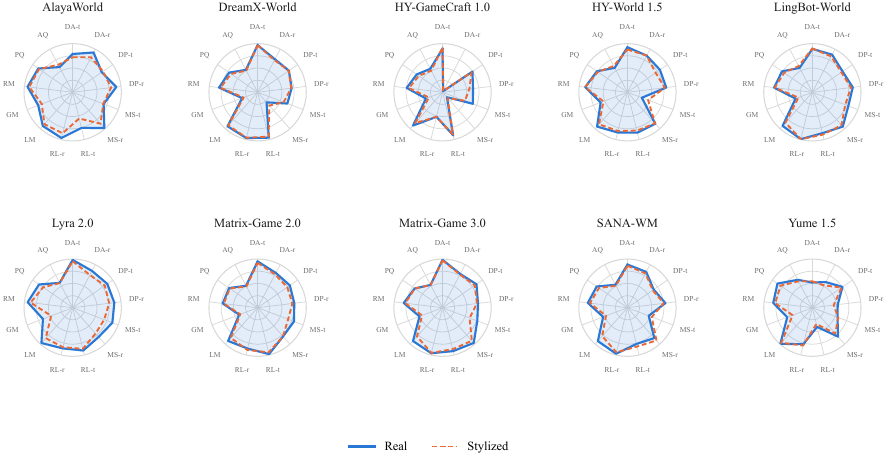}
  \caption{Capability profile of each model over the thirteen reported columns,
           with First-Person Real (solid) and First-Person Stylized (dashed)
           overlaid. All axes run $0$ to $100$ in the same order in every panel,
           so panels are directly comparable and the gap between the two
           outlines is that model's stylization penalty. Read the \emph{shapes},
           not the areas: a balanced polygon and a deeply notched one can
           enclose similar area while being very different systems to control.
           That the two outlines nearly coincide is itself the finding ---
           stylization moves these scores by $5.1$ points on average --- and the
           systematic part of that shift is visible only on the three memory
           axes, where every model moves the same way.}
  \label{fig:supp_profiles}
\end{figure}

\figref{fig:supp_profiles} plots every model on both first-person splits. Two things
the numbers do not show.

First, there is one all-round model and no second. Lyra~2.0 is the only
near-convex profile, at or above the suite mean on 12 of 13 columns; HY-World~1.5
reaches 10 and no other model exceeds 9. Every remaining profile has at least
four notches, so the suite has no crowd of near-equal generalists whose ranking
would hinge on weighting.

Second, the stylization penalty falls hardest on the strongest models. Summed
over the thirteen columns it ranges from $38.8$ for DreamX-World to $110.6$ for
Lyra~2.0 and $98.3$ for AlayaWorld --- precisely the two models that lead on
Real. Robustness to style is therefore a distinct axis of merit rather
than a corollary of performance.


\section{Qualitative Examples per Metric}
\label{supp:qualitative}

\begin{figure}[t]
  \centering
  \includegraphics[width=\textwidth]{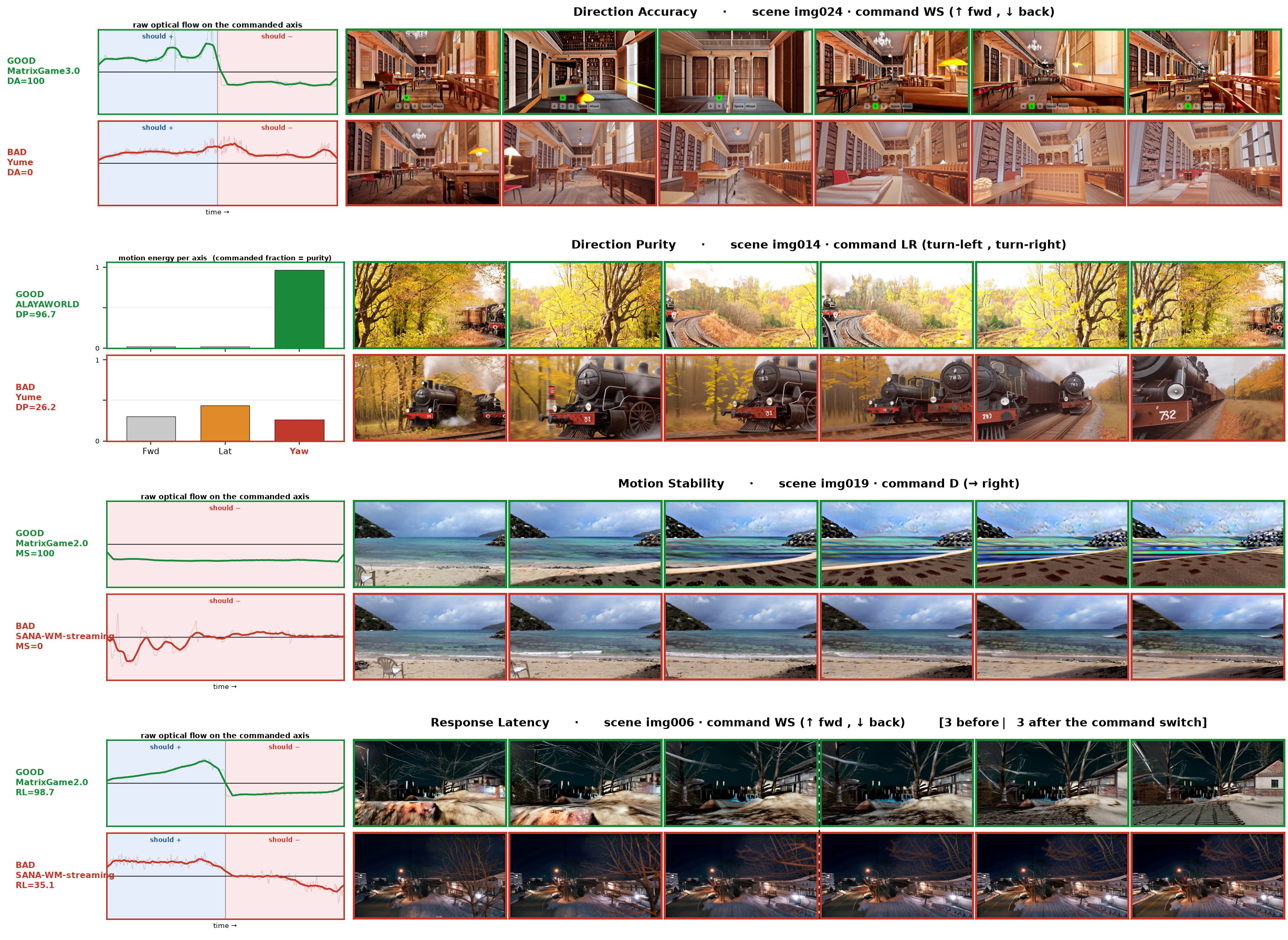}
  \caption{One high- and one low-scoring case for each Action Dynamics metric.
           Every row is one model on one commanded sequence: the panel at the
           left is the signal the metric reads, the six frames at the right are
           sampled from the same segment. Green marks the high-scoring model, red
           the low-scoring one, with the value beside it. Scene, command, and
           sampling are fixed within each pair, so the only difference is the
           model. Shading in the trace panels marks what the flow on the
           commanded axis \emph{should} do; the dashed rule in Response Latency
           is the command switch. Row labels abbreviate the released checkpoint
           names of \tabref{tab:supp_checkpoints}.}
  \label{fig:supp_examples_action}
\end{figure}

\begin{figure}[t]
  \centering
  \includegraphics[width=\textwidth]{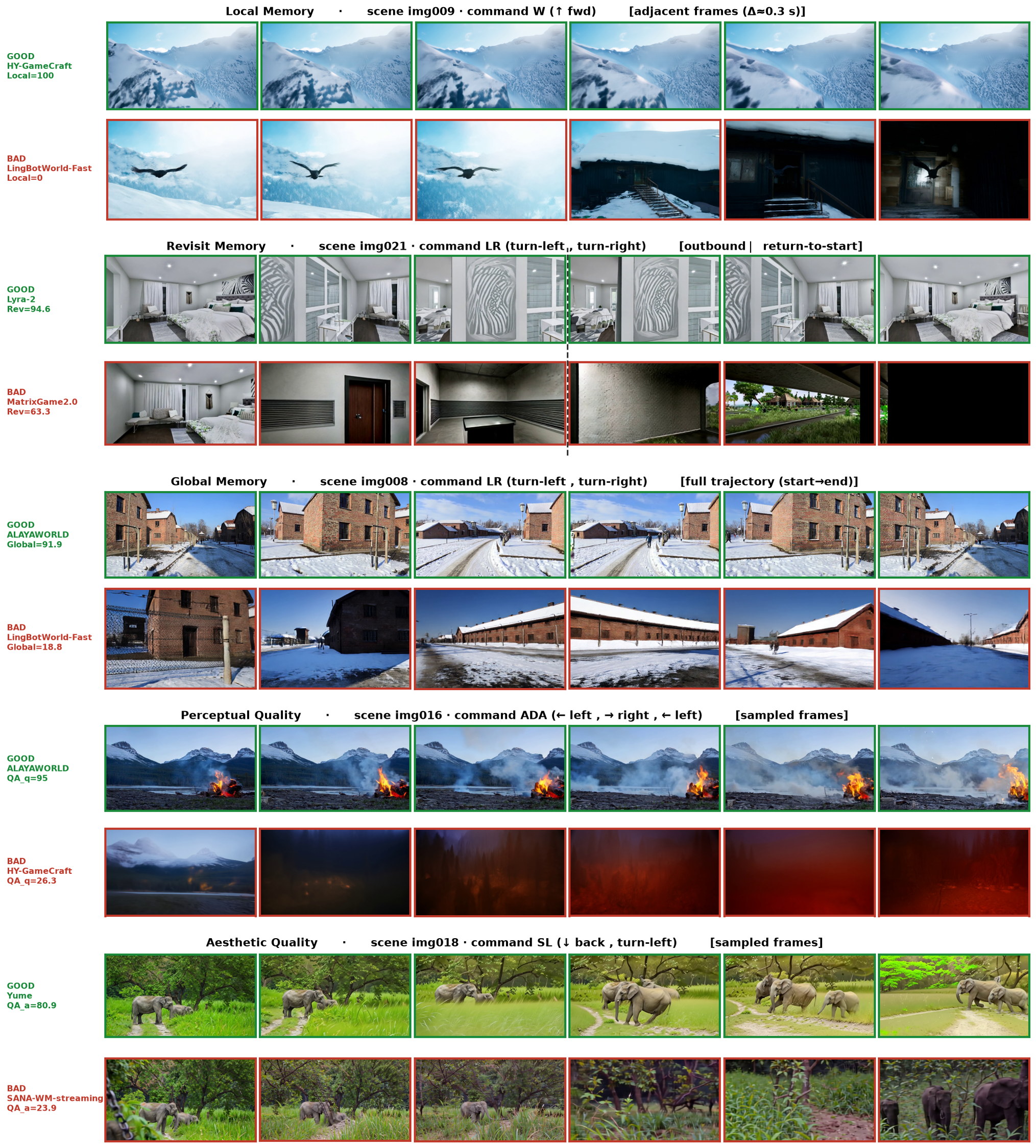}
  \caption{The same treatment for the metrics that read the whole video rather
           than its motion axes. Frames are drawn from the span each metric
           actually reads: adjacent frames for Local Memory, the outbound and
           return halves either side of the dashed rule for Revisit Memory, and
           the full trajectory otherwise. No trace panel is needed here, because
           these failures are visible in the frames themselves. Row labels
           abbreviate the checkpoint names of \tabref{tab:supp_checkpoints}.}
  \label{fig:supp_examples_rest}
\end{figure}

\figref{fig:supp_examples_action} and \figref{fig:supp_examples_rest} pair a
high- and a low-scoring model for each of the nine metrics, holding scene and
command fixed within each pair.

\paragraph{Why the action panels differ from each other.} The Action Dynamics
rows carry a signal panel as well as frames, because their failures are in the
motion rather than in the pixels: six stills cannot show that a model kept
turning after the command changed. Three of the four then read the same
quantity --- the optical flow on the commanded axis as a function of time --- and
differ only in what they ask of it: whether its sign matches the command
(Accuracy), whether it holds steady (Stability), and how soon it changes when the
command does (Latency). Direction Purity asks something no single channel can
answer, namely how much of the total motion is on the commanded axis at all, so
its panel is an energy split across the three channels and the commanded
channel's share is the score. That those bars carry no sign is the property
\secref{supp:da_dp} turns on: Purity is a ratio of magnitudes.

\paragraph{What the failures look like.} Direction Accuracy falls to $0$ where
the trace never crosses the axis, the model advancing through both halves of a
forward-then-backward command. Direction Purity falls where a model told to turn
puts more motion into strafing than into yaw, leaving the commanded channel not
even the largest of the three bars. Motion Stability falls to $0$ on a trace that
oscillates and then decays to nothing while the command still runs, and Response
Latency separates a step at the switch from a slow ramp that arrives seconds
late. Among the remaining metrics, Local Memory falls to $0$ where the sequence
cuts outright, mid-command, from an open mountainside to a building interior.
Revisit Memory separates a model that returns to the room it left from one that
arrives somewhere else entirely after the same pair of turns. Global Memory
distinguishes a scene whose layout survives a full sweep from one whose
structures rearrange as the camera passes, which no pair of adjacent frames would
reveal. The two quality metrics fall as their names suggest, their low-scoring
rows degrading into haze and clutter while the commanded motion still runs.


\end{document}